# BAYESIAN NON-PARAMETRIC MODEL TO TARGET GAMIFICATION NOTIFICATIONS USING BIG DATA

Meisam Hejazi Nia
Brian Ratchford

Naveen Jindal School of Management, Department of Marketing Science, SM32

The University of Texas at Dallas

800 W. Campbell Road

Richardson, TX, 75080-3021

## INTRODUCTION

User generated content (UGC) is the cornerstone of social and online marketing. However, the key challenge for online marketers to leverage UGC is to encourage users to generate more quality content. To overcome this burden, practitioners have started to use video game concepts such as badges, leaderboard, and points to encourage users, under the umbrella of an approach called Gamification. Online marketers require a data driven approach to target users based on their response to gamification elements. Knowing the response of individual users to various game elements can help the online marketer to emphasize various content generating tasks in its personal messaging, to maximize the total number of user generated contents. For example, knowing that a user reduces its content contribution after receiving a badge, an online marketer can create a diversified list of content generating tasks for user in a customized message, to make badge earning more difficult. Moreover, knowing that a user increases its content contribution after earning more points, the online marketer can create a targeted list of content generating tasks for users in a customized message, to make badge earning simpler.

Online marketers can leverage their massive data sets of users' content generations to create more customized targeted messages. This big data usually consists of several little data sets for each user, but its key advantage relative to the classic data sets is that it has more information about the tail of the distribution of customer response. This tail is relevant for targeting. Of course, a model can accommodate capturing the behavior on tail, if it allows the number of parameters to grow with the size of the data set. A useful method shall not through away these data by sampling, but it shall be flexible to not to misfit.

Hierarchical Bayesian (HB) approaches are well known for their estimation of individual specific parameters, and for allowing for unobserved heterogeneity, while sharing statistical strength across individual parameters. However, to be flexible, an HB model shall deviate from the normal prior on the consumer response parameters to the mixture normal structure, to capture behavior parameter of users in tail. Furthermore, a suitable method for Big Data shall be not only scalable, but also fast, to allow an online marketer to target its users in timely manner. In summary, a suitable approach shall create a computationally tractable solution for the computationally hard gamified targeting problem for big data.

The current proposed model uses hierarchical Bayesian sparse modeling approach for users' content generating choices to allow for users' unobserved heterogeneity. It exercises a mixed logit model, with individual specific random effects that control for self-selection. To address scalability and flexibility concerns, I used a version of stochastic optimization approach called mini-batch gradient descend. Unlike the batch approach that uses complete data set to update the parameters, the mini-batch approach iteratively and randomly samples data to create a noisy measure of gradient and hessian of the objective function. Studies show that under regularity conditions the mini-batch approach can converge to the batch optimization approach. However, the advantage of the mini-batch approach is that it uses less memory, and it is computationally faster. In addition, the proposed approach estimates the mixed logit model in two steps. In the first step, it uses the observed data to identify the segment membership of each user. The BIC measure identifies the number of segments. Then, in the second step, conditional on the segment membership the model, it optimizes a-posteriori of the parameters. In summery, the current approach sets the number of segments exogenously, using BIC measure.

Although the mentioned approach is not wrong, a better approach involves endogenizing the number of segments. A realistic approach should not even assume the number of segments, rather it shall assume that the world is infinitely complex, so it shall allow the model to automatically select the finite number of segments observed in the finite data set. This way the approach can be general enough to update the number of segments as firm observes more data. As a result, the learning of an online marketer from its big data is not limited anymore, and the marketer learns more about its users, as it observes more data. In fact a good approach should allow the firm to update its segmentation based on latent information set that it has captured from the streaming data. This segmentation might also evolve across time as users' latent motivation state changes. Therefore, an online marketer requires a dynamic segmentation technique. This way the online marketers' posterior belief about parameters evolves, as the marketer updates its belief conditioning on the latest information.

In fact, big data makes offline model selection computationally intractable because estimating a non-linear model over a big data for a specific model structure is time consuming. Non-parametric Bayesian provides tools for this computationally hard automatic model structure selection problem. The new approach I have planned to use falls into non-parametric Bayesian approaches category. In particular, to model users' latent motivation I use infinite Hidden Markov Model (iHMM). In this approach, I assume that users have various latent motivation states that are time varying. These latent time varying motivation states define users' response parameters to gamification elements, in choosing whether to contribute content or not. In this structure, the transition probability between states is modeled as a Hierarchical Dirichlet Process (HDP), and the emission probability is modeled as ordered logit model of users' content contribution choice given the latent state motivation and the users' gamification earnings (i.e. badges, rank on the leaderboard, reputation points).

To control for unobserved heterogeneity across users, further I use a Dirichlet Process (DP) on the parameters of the ordered logit emission probability model. As a result the model has two building blocks of iHMM and DP to allow automatic model structure selection over big data, by endogenizing the number of user segments and states. These approaches are scalable, flexible, realistic, and machine learning literature shows that they improve prediction; however, their estimation with MCMC method suffers from slow convergence, and slow mixing problem. Therefore, to allow an online marketer to learn parameter of users responses in timely manner to target them, conditional on the latest information, I use a combination of Particle Learning (PL) and Variational Bayesian (VB). These approaches help to speed up the estimation. To estimate the iHMM model, I will use PL. PL is a Sequential Monte Carlo (SMC) method that uses simulation based on discrete approximation of a random cloud of particle to estimate the targeted posterior density. Its advantage is that it allows belief updating over parameters based on the latest observed information set in a computationally tractable way. I parallelize the PL process to speed up estimation. To speed up estimation of the DP over users specific parameters, I use a Variational Bayesian (VB) approach. This approach maximizes the evidence lower bound for the K-L divergence of parameters to approximate the parameters of the factorized variational distribution of user specific parameters.

All in all, I suggest an approach that helps the online marketers to target their gamification elements to users by modifying the order of the list of tasks that they send to users. It is more realistic and flexible as it allows the model to learn more parameters when the online marketers collect more data. The targeting approach is scalable and quick, and it can be used over streaming data.

## MODEL

The proposed model has two non-parametric Bayesian building blocks: Infinite Hidden Markov Model and Bayesian infinite Gaussian mixture model. The former is analogous to Chinese Franchise process (CFP) or Hierarchical Dirichlet Process (HDP), and the latter is analogous to Chinese restaurant process (CRP), or Dirichlet process (DP). CRP describes how customers entering a restaurant might select a table with more customers with higher probability and a new empty table with tiny probability. CFP describes a dynamic process in which there are tourists that enter a restaurant in the first night of their trip according to CRP, but the list of restaurants to visit for next night is defined per tables.

For the first building block, the Infinite Hidden Markov Model in our case explains the probability of answering to a questions with a vector of gamification assets that user has accumulated until time t (including number of questions answered, number of answered received, total reputation points, reputation points earned last week, last week rank on leaderboard, first order difference rank on leader board, total gold, silver, and bronze badges earned and tags attached to them, and gold, silver, and bronze badges earned a moment ago and tags attached to them), based on a parameter that varies based on unobserved motivation of user i at time t.

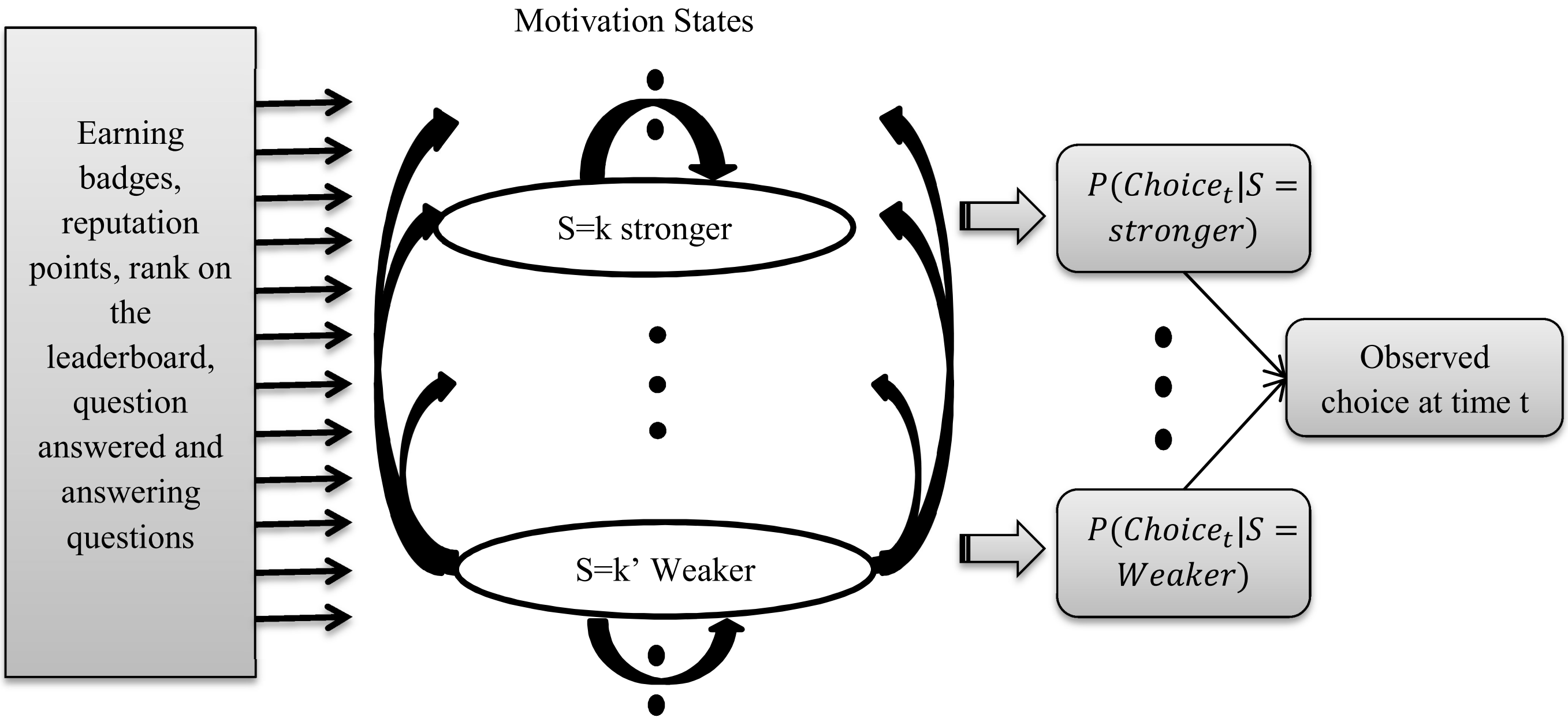

We call the probability of this choice emission probability, and the probability of user's transition from one state of motivation to another, the transition probability, consistent with the terminology of Hidden Markov Model. Formally, the transition probability has the following form:

$$\Pi_{i,t-1\rightarrow t} = \begin{array}{c} t-1 \\ \vdots \\ k \\ k+1 \\ \vdots \end{array} \begin{array}{c} \cdots \quad k \quad k+1 \quad \cdots \\ \begin{bmatrix} \vdots & \vdots & \vdots & \vdots \\ \cdots & \pi_{ikk} & \pi_{i,k,k+1} & \cdots \\ \cdots & \pi_{i,k+1,k} & \pi_{i,k+1,k+1} & \cdots \\ \vdots & \vdots & \vdots & \vdots \end{bmatrix} \end{array}$$

The infinite Hidden Markov generative process is formally defined as follows:

$$\begin{aligned}
&\lambda_{i\tau} \sim Beta(1,\kappa) \\
&\beta_i \sim STB(\lambda_i) \equiv \beta_i = \lambda_{i\tau} \prod\nolimits_{t=1}^{\tau-1} (1-\lambda_{it}) \\
&\alpha_i \sim Gamma(a_0, b_0) \\
&\pi_{ik} \sim DP(\alpha_i, \beta_i) \\
&s_{it} \sim Multinomial(\pi_{is_{it-1}}) \\
&\Gamma_{il} \sim H \equiv \Gamma_{il} \sim N(\mu_{\Gamma il}, \sigma^2_{\Gamma il}), \Lambda_{il} = (\mu_{\Gamma il}, \log(\sigma^2_{\Gamma il})), l = 1,\ldots,\infty \\
&y_{it} \sim Logit(\Gamma_{is_{it}})
\end{aligned}$$

where

$\lambda_{i\tau}$ denotes the hyperparameter that controls the number of states, and it has beta distribution with parameter $\kappa$.

$\beta_i$ denotes the prior distribution on each row of the transition matrix, which has stick breaking construction, with parameter $\lambda_\tau$.

$\alpha_i$ denotes the concentration (or novelty) parameter for the Dirichlet process rows of transition matrix. It controls how similar rows of the transition matrix are to each other, or how sparse are the rows. This parameter has Gamma distribution with parameters $a_0$ and $b_0$.

$\pi_{ik}$ denotes the k'th rowh of the transition matrix of user k, which has Dirichlet process distribution, with concentration parameter $\alpha_i$ and baseline distribution $\beta_i$.

$s_{it}$ denotes the motivation state of user i at time t, which is an indicator with multinomial distribution that has a $\pi_{is_{it-1}}$ state specific $s_{it-1}$ parameter corresponding to the row in the transition matrix of state in previous moment $s_{it-1}$

$\Gamma_{il}$ denotes emission (distribution of observed choice) parameter in the logistic link function that connects unobserved motivation state $l$ with observed choice of answering a question. This parameter has normal distribution with mean $\mu_{\Gamma i}$ and variance $\sigma^2_{\Gamma i}$.

$\Lambda_i = (\mu_{\Gamma i}, \log(\sigma^2_{\Gamma i}))$ denotes individual specific vector of mean and log variance of emission parameter distribution.

$y_{it}$ denotes the observed choice of user i at time t, indicating whether user answered a question or not. The probability of this choice follows logistic distribution with parameter $\Gamma_{s_{it}}$.

The model allows for accounting for new motivation states that can be lower or higher in the future. The distribution of choice parameters across the user population summarizes the population behavior given each motivation state.

State dependent choice model:

$$U_{i|s_{it}} = \gamma^{i}_{i|s_{it}} + \gamma^{t}_{i|s_{it}} + \gamma^{1}_{i|s_{it}} cont_{it-1} + \gamma^{2}_{i|s_{it}} rcv_{it-1} + \gamma^{3}_{i|s_{it}} crep_{iw-1} + \gamma^{4}_{i|s_{it}} rep_{iw-1} + \\ \gamma^{5}_{i|s_{it}} rnk_{iw-1} + \gamma^{6}_{i|s_{it}} \Delta rnk_{iw-1} + \gamma^{7}_{i|s_{it}} bdg_{it-1} + \gamma^{8}_{i|s_{it}} tag_{it-1} + \\ \gamma^{9}_{i|s_{it}} cbdg_{it-1} + \gamma^{10}_{i|s_{it}} ctag_{it-1} + \varepsilon_{i|s_{it}}$$

where

$U_{i|s_{it}}$ is the utility of user i to answer new questions given latent motivation state and gamification assets,

$\Gamma_{s_{it}} = (\gamma^{i}_{i|s_{it}}, \gamma^{t}_{i|s_{it}}, \gamma^{1}_{i|s_{it}}, .., \gamma^{10}_{i|s_{it}})$ is the state-specific parameter of response, while the first two parameters are controlling individual specific and time specific random effects, and

$x_{it} = (cont_{it-1}, rcv_{it-1}, crep_{iw-1}, rep_{iw-1}, rnk_{iw-1}, \Delta rnk_{iw-1}, bdg_{it-1}, tag_{it-1}, cbdg_{it-1}, ctag_{it-1})$ is a vector of time-varying covariates associated with the gamification assets (i.e. number of questions answered, number of answered received, total reputation points, reputation points earned last week, last week rank on leaderboard, first order difference rank on leader board, total gold, silver, and bronze badges earned and tags attached to them, and gold, silver, and bronze badges earned a moment ago and tags attached to them) of user i at time t, and user choice. Motivation for this utility structure is presented at the end of this paper.

$$P_i(Y_{it} = y_{it} \mid S_{it} = s_{it}) = m_{i|s_{it}} = \frac{\exp(U_{i|s_{it}})}{1 + \exp(U_{i|s_{it}})}, s \in \{1, .., \infty\}$$

In summary, the iHMM models the observed responding behavior of users as a noisy signal of hidden motivation state. This unobserved motivation state evolves with stochastic process dynamically. The form of the stochastic process is first order Markov. Another interpretation of this process is that from econometrician point of view users are segmented based on their unobserved motivation state dynamically. The flexible structure of transition matrix allows iHMM model to capture any type of dynamic that can be assumed for users' transition between motivation states.

The second building block controls for unobserved heterogeneity in users response parameter. This building block consists of Bayesian Dirichlet process prior on the emission parameters of choice given state of motivation. Formally, this generative process's structure is defined as follows:

$$\alpha_{v0} \sim Gamma(a_{v0}, b_{v0})$$
$$v_k \sim Beta(1, \alpha_{v0})$$
$$\theta_k \sim STB(V) \equiv \theta_k = v_k \prod_{j=1}^{k-1}(1-v_j) \equiv \theta_1,...,\theta_K \sim Dir(\alpha_0 / K,...,\alpha_0 / K), K \to \infty$$
$$G_0 \equiv \quad N(\mu_{G0}, \Sigma_{G0}) IW(\vartheta_{G0}, S_{G0})$$
$$G \sim DP(\alpha_{v0}, G_0), \eta_i \sim G \equiv G(\eta_i) = \sum_{k=1}^{\infty} \theta_k \delta(\eta_i, \eta_k^*) \equiv \eta_i = \begin{cases} \eta_k^* \text{ with prob } \dfrac{n_i}{I-1+\alpha_{v0}} \\ \eta, \eta \sim G_0 \text{ with prob } \dfrac{\alpha_{v0}}{I-1+\alpha_0} \alpha_{v0} \end{cases}$$
$$c_{il} \sim Mult(\theta) \equiv c_{il} \sim Discrete(\theta_1,...,\theta_K), K \to \infty$$
$$\Lambda_{il} \mid c_{il} \sim F(.;\eta_{c_{il}}) \equiv p(\Lambda_{il} \mid c_{il}) \sim N(\Delta D_i + \mu_{\Lambda c_{il}}, \Sigma_{\Lambda c_{il}}) \equiv p(\Lambda_{il}) = \sum_{k=1}^{\infty} \theta_k N(\Delta D_i + \mu_{\Lambda c_{il}}, \Sigma_{\Lambda c_{il}})$$
$$\Delta \sim N(\mu_\Delta, \sigma_\Delta^2)$$

where

$\alpha_{v0}$ denotes the distribution of concentration (or novelty) parameter of Dirichlet process, which controls how similar are the partitions (i.e. user segments).

$v_k$ denotes the parameter that controls number of partitions (i.e. number of user segments), and it has beta distribution.

$\theta_k$ denotes mixture probability/proportion (latent probability measure), and it has stick breaking process construction.

$G_0$ denotes the distribution for the prior on each partition's mean and precision, which has conjugate normal inverse Wishart structure.

$\eta_i$ denotes mean and precision parameter of each partition (i.e. users segment), which has Dirichlet Process distribution denoted by G.

$c_{il}$ denotes the latent indicator of partition membership for user I, which has discrete multinomial distribution, with probability vector of $\theta$.

$\Lambda_{il}$ denotes the vector of emission (logistic) parameter distribution mean and log variance for user i that is in partition $c_{il}$ and in state l.

$D_i$ denotes the vector of demographics of user i.

$\Delta$ denotes the parameter to explain the effect of demographic on user's motivation to contribute across population. It has mean $\mu_\Delta$ and variance $\sigma_\Delta^2$.

The Dirichlet process allows for atomic distribution of the emission parameters. Per definition, a finite subset of random measures distributed with Dirichlet Process has Dirichlet distribution. Dirichlet distribution represents the distribution of random probability measures over a simplex, and Dirichlet process represents the distribution of random partition/assignment. The key property of Dirichlet process that allows its closure under marginalization is exchangeability of partitions and assignments. This property allows to the estimation procedure to use De Finnitti theorem to marginalize out the random measure, allowing for close form probability for assignment of given data point given the assignment of all other data points.

**ESTIMATION**

Probability of transition from one period to another is defined as:

$$P_i((Y_{it} = y_{it}, S_{it} = s_{it}) \mid (Y_{it-1} = y_{it-1}, S_{it-1} = s_{it-1})) = P_i(Y_{it} = y_{it} \mid S_{it} = s_{it}) P(S_{it-1} = s_{it-1}, S_{it} = s_{it})$$

Likelihood of an observed sequence of choices:

$$P_i(Y_{i1}=y_{i1},...,Y_{iT}=y_{iT})=\sum_{s_{i1}=1}^{\infty}\sum_{s_{i2}=1}^{\infty}\cdots\sum_{s_{iT}=1}^{\infty}\left[P(S_{i1}=s_{i1})\prod_{t=2}^{T}P(S_{it}=s_{it}\mid S_{it-1}=s_{it-1})\right]$$
$$\prod_{t=1}^{T}P(Y_{it}=y_{it}\mid S_{it}=s_{it})$$

Concretely the likelihood is:

$$P_i(Y_{i1}=y_{i1},...,Y_{iT}=y_{iT})=\sum_{s_{i1}=1}^{\infty}\sum_{s_{i2}=1}^{\infty}\cdots\sum_{s_{iT}=1}^{\infty}\left[\pi_{is_1}\prod_{t=2}^{T}q_{is_ts_{t-1}}\right]\prod_{t=1}^{T}m_{i|s_{it}}{}^{y_{it}}(1-m_{i|s_{it}})^{(1-y_{it})}$$

There are three ways to estimate the model: collapsed Gibbs sampler, Beam sampler, and Particle Filter. The former two methods are not suitable for online streaming data. Collapsed Gibbs sampler iteratively samples latent state by computing the probability of $p(s_{it}=k\mid s_{i,-t},\beta_i,\alpha_i,H)\propto p(y_{it}\mid s_{it},s_{i,-t},\beta_i,H)p(s_{it}=k\mid s_{i,-t},\beta_i,\alpha_i)$. The first factor is the integrated likelihood of observation given latent state and prior distribution on the parameter H, so it is: $H:\int p(y_{it}\mid s_{it},\Gamma_{s_{it}})p(\Gamma_{s_{it}}|s_{i,-t},y_{-t},H)d\Gamma_{s_{it}}$. Given that emission distribution and prior distribution on its parameter H are conjugate, this probability is easy to compute. Furthermore, the probability of each user i in a given motivation state at time t given all other times can be written as:

$$p(s_{it}=k\mid s_{i,-t},\beta_i,\alpha_i)\propto\begin{cases}(n_{s_{it-1},k_i}+\alpha_i\beta_{k_i})\dfrac{n_{k_i,s_{it+1}}+\alpha_i\beta_{s_{it+1}}}{n_{k_i\cdot}+\alpha_i} \; if \quad k_i\le K_i,\quad k\ne s_{i,t-1}\\ (n_{s_{it-1},k_i}+\alpha_i\beta_{k_i})\dfrac{n_{k_i,s_{it+1}}+1+\alpha_i\beta_{s_{it+1}}}{n_{k_i\cdot}+1+\alpha_i} \; if \quad k_i\le K_i,\quad k=s_{i,t-1}=s_{i,t+1}\\ (n_{s_{it-1},k_i}+\alpha_i\beta_{k_i})\dfrac{n_{k_i,s_{it+1}}+\alpha_i\beta_{s_{it+1}}}{n_{k_i\cdot}+1+\alpha_i} \; if \quad k_i\le K_i,\quad k=s_{i,t-1}\ne s_{i,t+1}\\ \alpha_i\beta_{k_i}\beta_{s_{it+1}} \; if \quad k_i=K_i+1\end{cases}$$

where

$n_{i,j}$ denotes the number of transitions from state i to state j, excluding time steps t-1 and t

$n_{.i}$ denotes the number of transitions into state i, excluding time steps t-1 and t

$n_{i.}$ denotes the number of transitions out of state i, excluding time steps t-1 and t

$s_{i,-t}$ denotes state of user i at each point in time, excluding time steps t-1 and t

$K_i$ denotes number of distinct states in $s_{i,-t}$

The Gibbs sampler is useful as it is straight forward. However, it suffers from one major drawback: sequential and time series data are likely to be strongly correlated, which slows up mixing and convergence. The Beam sampler does not suffer from this slow mixing behavior, as it samples the whole sequence of states s in one go, through slice sampling and an auxiliary variable, but it is not useful for online streaming data. Particle Learning (PL) however is suitable approach for updating parameters based on the last information observed. Particle learning is an application of Sequential Monte Carlo (SMC) sampler. SMC sequentially updates the distribution once a new observation is collected. PL particle approximation for state $s_{i,t}$ and the structural parameters $\zeta_i = (\widehat{\alpha}_{it}^{(b)}, \{\widehat{\beta}_{lit}^{(b)}\}, \widehat{\lambda}_{it}^{(b)}, \{\widehat{\Gamma}_{ilt}^{(b)}\})$, as follows:

$$\widehat{p}(s_{i,t}, \zeta_i \mid y_{i1}, \ldots, y_{it}) = \sum_{b=1}^{B} w_t^{(b)} \delta_{(\widehat{s}_{i,t}, \widehat{\zeta}_i)}(s_{i,t}, \zeta_i)$$

where

$\delta_{(\widehat{s}_{i,t}, \widehat{\zeta}_i)}(s_{i,t}, \zeta_i)$ denotes Kronecker (Dirac) delta function that represents a pulse function (a function that is zero everywhere except at subscript, at which it is one).

$B$ denotes the number of particles used for approximation.

PL makes two assumptions:

First, at any time t, the posterior distribution for structural parameter $\zeta_i$ depends on the states and observations through a low dimensional vector of sufficient statistics $r_{it}$, which can be sequentially updated using recursion $R$ such that $r_{i,t+1} = R(r_{it}, y_{i,t+1}, s_{i,t+1})$, so that $p(\zeta_i \mid s_{i,1},...,s_{i,t}, y_{i,1},..., y_{i,t}) = p(\zeta_i \mid r_{i,t})$. This way the system should only keep track of the sufficient statistics rather than the history of motivation states and observations.

Second, PL requires that the predictive distribution $p(y_{i,t} \mid r_{i,t}, s_{i,t}, \zeta_i) = \int\int p(y_{i,t} \mid r_{i,t+1}, s_{i,t+1}) p(r_{i,t+1}, s_{i,t+1} \mid r_t, s_t, \zeta_i) dr_{i,t+1} ds_{i,t+1}$ can be computed in closed form. If these two conditions are satisfied, we can treat the sufficient statistics $r_t$ as deterministically updated state, and write:

$$p(r_{i,t}, s_{i,t}, \zeta_i \mid y_{i1},.., y_{i,t+1}) \propto p(y_{i,t+1} \mid r_{i,t}, s_{i,t}) p(r_{i,t}, s_{i,t} \mid y_{i1},.., y_{i,t})$$

As a result

$$\begin{aligned} p(r_{i,t+1}, s_{i,t+1}, \zeta_i \mid y_{i1},.., y_{i,t+1}) = \int\int\int & p(\zeta_i \mid r_{i,t+1}) p(r_{i,t+1} \mid r_{i,t}, \zeta_i^*, y_{i,t+1}) p(s_{i,t+1} \mid s_{i,t}, \zeta_i, y_{i,t+1}) \\ & p(r_{i,t}, s_{i,t}, \zeta_i^* \mid y_{i1},.., y_{i,t+1}) d\zeta_i^* dr_{i,t} ds_{i,t} \end{aligned}$$

where $p(r_{i,t+1} \mid r_{i,t}, \zeta_i^*, y_{i,t+1})$ is a point mass concentrated in $R(r_{it}, y_{i,t+1}, s_{i,t+1})$.

Logistic specification of emission link function does not satisfy the second condition. To satisfy this condition we need a conjugate distribution for the distribution of choice parameters. The data augmentation structure suggests the following form:

$$\begin{aligned} & y_{it} = I(U_{i|s_{it}} \geq 0) \\ & U_{i|s_{it}} = \Gamma_{s_{it}} x_{it} + \varepsilon_{i|s_{it}} \quad where \quad \varepsilon_{i|s_{it}} \sim -\ln \mathrm{E}(1) \end{aligned}$$

where

$\varepsilon_{i|s_{it}}$ is an extreme value distribution of type 1.

$\mathrm{E}(1)$ is an exponential of mean one.

$I(.)$ is an indicator function that is equal to one when the condition is met.

Frunwirth-Schnatter and Frunwirth-Schnatter (2007) suggest a 10-component mixture of normal distribution to approximate extreme value distribution. This structure satisfies the conjugacy condition. In this structure, conditional on the indicator of the component $z_{i|s_{it}}$, the model formally becomes:

$$\begin{aligned} y_{it} &= I(U_{i|s_{it}} \geq 0) \\ U_{i|s_{it}} &= \Gamma_{s_{it}} x_{it} + \varepsilon_{i|s_{it}} \; where \quad \varepsilon_{i|s_{it}} \sim N(\mu_{z_{i|s_{it}}}, \sigma_{z_{i|s_{it}}}), P(z_{i|s_{it}} = j) = w_j \end{aligned}$$

Given $z_{i|s_{it}}$, we have conditional sufficient statistics recursion for $\Gamma_{s_{it}}$ as follows:

$$r_{t+1}^{\Gamma_{s_{it}}} = R(r_t^{\Gamma_{s_{it}}}, U_{i|s_{it}}^{t+1}, z_{i|s_{it}}^{t+1}, y_{it+1}, \eta_{c_i}, c_i)$$

Concretely, the sufficient statistics includes the following vector:

$$r_t^{\Gamma_{s_{it}}} = (\bar{x}_i, \overline{U}_{i|s_{it}}, C_{xx}, C_{xU_{i|s_{it}}}, C_{U_{i|s_{it}}U_{i|s_{it}}})$$

Where

$\bar{x}, \overline{U}_{i|s_{it}}$ are mean of the gamification asset vector, and latent utility given the component membership, and

$C_{xx}, C_{xU_{i|s_{it}}}, C_{U_{i|s_{it}}U_{i|s_{it}}}$ are the variances and covariance of the gamification asset vector and latent utility given the component membership.

Therefore, the recursion for each of the sufficient statistics has the following form:

$$\begin{aligned}
\overline{x}_i^{(t+1)} &= \overline{x}_i^{(t)} + \frac{1}{t+1}(x_{it+1} - \overline{x}_i^{(t)}) \\
\overline{U}_{i|s_{it}}^{(t+1)} &= \overline{U}_{i|s_{it}}^{(t)} + \frac{1}{t+1}(U_{i|s_{it+1}}^{t+1} - \overline{U}_{i|s_{it}}^{(t)}) \\
C_{ixx}^{(t+1)} &= \frac{t}{t+1}(C_{ixx}^{(t)} + (\overline{x}_i^{(t)})^2) + (\frac{x_{it+1}^2}{t+1}) - (\overline{x}_i^{(t+1)})^2 \\
C_{xU_{i|s_{it}}}^{t+1} &= \frac{t}{t+1}(C_{xU_{i|s_{it}}}^{(t)} + (\overline{x}_i^{(t)}\overline{U}_{i|s_{it}}^{(t)})) + (\frac{x_{it+1}U_{i|s_{it+1}}^{t+1}}{t+1}) - (\overline{x}_i^{(t+1)}\overline{U}_{i|s_{it}}^{(t+1)}) \\
C_{U_{i|s_{it}}U_{i|s_{it}}}^{(t+1)} &= \frac{t}{t+1}(C_{U_{i|s_{it}}U_{i|s_{it}}}^{(t)} + (\overline{U}_{i|s_{it}}^{(t)})^2) + \frac{(U_{i|s_{it+1}}^{t+1})^2}{t+1} - (\overline{U}_{i|s_{it}}^{(t+1)})^2
\end{aligned}$$

It is important to note that the sufficient statistics is state specific, so for each of the hidden states these statistics should be tracked. In other word, the drawn latent state indicator defines the sufficient statistics of which latent state should be updated for each particle.

Frunwirth-Schnatter and Frunwirth-Schnatter (2007) furthermore computes the parameters of the 10 component mixture model as follows:

| z | 1 | 2 | 3 | 4 | 5 | 6 | 7 | 8 | 9 | 10 |
|---|---|---|---|---|---|---|---|---|---|---|
| $w_z$ | 0.00397 | 0.0396 | 0.168 | 0.147 | 0.125 | 0.101 | 0.104 | 0.116 | 0.107 | 0.088 |
| $\mu_z$ | 5.09 | 3.29 | 1.82 | 1.24 | 0.764 | 0.391 | 0.0431 | -0.306 | -0.673 | -1.06 |
| $\sigma_z$ | 4.50 | 2.02 | 1.10 | 0.422 | 0.198 | 0.107 | 0.0778 | 0.0766 | 0.0947 | 0.146 |

These expressions lead to the following iterative algorithm to update the filtering distribution:

**Algorithm1** Particle Learning Filtering

Sample $\hat{\zeta}_{i0}^{(b)} \sim p(\zeta_i)$ and $\hat{s}_{i0}^{(b)} \sim p(s_{i0} \mid \hat{\zeta}_{i0}^{(b)})$, and initialize $\hat{r}_{i0}^{(b)}$.

**For** t=0 to T-1 **do**:

**(step-ahead prediction)** Set $\vartheta_{it}^{(b)} = p(y_{it+1} \mid \widehat{s}_{it}^{(b)}, \widehat{r}_{it}^{(b)}, \widehat{\zeta}_{it}^{(b)})$

**(Re-sample)**: current particles $\left\{(\widetilde{s}_{it}, \widetilde{r}_{it}, \widetilde{\zeta}_{it})^{(b)}\right\}_{b=1}^{B}$ by generating an index $ind(b) \sim Multi(\omega^{(b)})$, where

$$\omega^{(b)} = \frac{\vartheta_t^{(b)}}{\sum_{b'=1}^{B} \vartheta_t^{(b')}}$$

**(Propagate)**: Sample

$$\hat{s}_{it+1}^{(b)} \sim p(s_{it+1}^{(b)} \mid \widetilde{s}_{it}^{(b)}, \widetilde{r}_{it}^{(b)}, \widetilde{\zeta}_{it}^{(b)}, y_{it})$$

**(Update)** : sufficient statistics

$$\widehat{r}_{i,t+1}^{(b)} = R(\widetilde{r}_{it}^{(b)}, y_{i,t+1}, \widetilde{s}_{i,t+1}^{(b)}, \widetilde{U}_{i|s_{it}}^{(b)})$$

**(Sample)**: draw structural parameters given their posterior distributions, conditioned on information available up to time t (MCMC adaptation)

$$\widehat{\zeta}_{i0}^{(b)} \sim p(\zeta_i \mid \widehat{r}_{i,t+1}^{(b)})$$

**End for**

Once filtering algorithm has been run for $t = 1,...,T$, the stored particles representation for the marginal distributions $\{p(r_{it}, s_{it}, \zeta_{it} \mid y_{it},..., y_{i1})\}_{t=1}^{T}$ can be used to generate sample paths from the joint distribution $p(r_{iT}, s_{iT},..., r_{i1}, s_{i1}, \zeta_{it} \mid y_{it},..., y_{i1})$ by using smoothing algorithm, repeated to generate B' sample paths, as follows:

**Algorithm2** Particle Learning Smoothing

Sample $(s_{iT}^{(b')}, r_{iT}^{(b')}, \zeta_i^{(b)'}) \sim p(s_{iT}, r_{iT}, \zeta_i \mid y_{i1},..., y_{iT}) = \sum_{b=1}^{B} \frac{1}{B} \delta_{(\widehat{s}_{iT}^{(b)}, \widehat{r}_{iT}^{(b)}, \widehat{\zeta}_{iT}^{(b)})}$

**For** t=T-1 to 1 **do**:

Set $q_{it}^{(b)} = p(s_{it}^{(b')}, r_{it}^{(b')} \mid \widehat{s}_{it}^{(b')}, \widehat{r}_{it}^{(b')}, \zeta_i^{(b')})$

**(Re-sample)** Set $\upsilon_{it}^{(b)} = \frac{q_{it}^{(b)}}{\sum_{b''=1}^{B} q_{it}^{(b'')}}$ and sample $(s_{it}^{(b')}, r_{it}^{(b')}) \sim \sum_{b''=1}^{B} \upsilon_{it}^{(b'')} \delta_{(\widehat{s}_{iT}^{(b'')}, \widehat{r}_{iT}^{(b'')})}$

**End For**

In order to write the particle learning algorithm for the infinite Hidden Markov Model consistent with Rodriguez (2011) an integrated likelihood for emission probability is required. This approach is less useful when there is uncertainty about the hyperparameters of the emission-probability's parameter-prior. In that case we have to draw particles on both hyperparameters and parameter of the emission-probability. I will discuss this aspect later. Formally, the integrated likelihood (which integrates out over the prior on the mean of regression parameter) has the following form for conditional exponential family:

$$y_i = x_i\beta + \varepsilon_i, \varepsilon_i \sim N(0,\sigma^2)$$

$$p(y_i \mid x_i, \beta, \sigma^2) = \frac{1}{\sqrt{2\pi\sigma^2}}\exp\{\frac{-(y_i - x_i\beta)^2}{2\sigma^2}\} = \frac{1}{\sqrt{2\pi\sigma^2}}\exp\{\frac{-y_i^2}{2\sigma^2} + \frac{y_i x_i \beta}{\sigma^2} - \frac{(x_i\beta)^2}{2\sigma^2}\}$$

$$p(y_i \mid \eta, x_i) = h_l(y_i, x_i)\exp\{\eta^T t(y_i, x_i) - a_l(\eta, x_i)\}$$

$$\eta = \langle \beta \rangle, t(y_i, x_i) = \left\langle \frac{y_i x_i}{\sigma^2} \right\rangle, a_l(\eta, x_i) = \frac{1}{2}(\frac{x_i^2}{\sigma^2}\beta^2), a_l(\eta) = (\beta^2), \log(h_l(y_i, x_i)) = \log(h_l(y_i)) = \frac{-y_i^2}{2\sigma^2} + \log\sigma^2$$

$$\beta \sim N(\beta_0, \tau_0^2) \Rightarrow p(\eta \mid \lambda) = h_c(\eta)\exp\{\lambda_1^T\eta + \lambda_2^T(-a_l(\eta)) + a_c(\lambda)\}, a_c(\lambda) = \tfrac{1}{2}(\beta_0 / \tau_0^2 + \log(\tau_0^2))$$

$$\begin{aligned} p(y_{1:n} \mid \lambda, x_i) &= \int (\textstyle\prod_{i=1}^n p(y_i \mid \eta, x_i)) p(\eta \mid \lambda) d\eta = \int (\textstyle\sum_{i=1}^n \log h_l(y_i) + \eta^T(\textstyle\sum_{i=1}^n t(y_i, x_i)) \\ &\quad - (\textstyle\sum_{i=1}^n x_i^2) a(\eta)\} \exp\{\lambda_1^T\eta + \lambda_2^T(-a_l(\eta, x_i)) + a_c(\lambda)\} d\eta = \\ &\quad \exp\{\textstyle\sum_{i=1}^n \log h_l(y_i)\} \int \exp\{\eta^T(\lambda_1^T + \textstyle\sum_{i=1}^n t(y_i, x_i)) - (\lambda_2^T + \textstyle\sum_{i=1}^n \frac{x_i^2}{\sigma^2}) a_l(\eta)\} d\eta \exp\{+a_c(\lambda)\} \end{aligned}$$

where the integral is un-normalized posterior distribution of hidden natural parameter, which is equal to the partition function of the posterior. In summary:

$$p(y_{1:n} \mid \lambda, x_i) = \exp\{\textstyle\sum_{i=1}^n h_l(y) + a(\hat{\lambda}) - a(\lambda)\}$$

where the posterior parameters have the following forms:

$$\beta \sim N(\beta_n, \tau_n^2)$$

$$\beta_n = \frac{\frac{\beta_0}{\tau_0^2} + \sum_{i=1}^{n} \frac{y_i x_i}{\sigma^2}}{\frac{1}{\tau_0^2} + \sum_{i=1}^{n} \frac{x_i x_i}{\sigma^2}}$$

$$\tau_n^2 = \frac{1}{\frac{1}{\tau_0^2} + \sum_{i=1}^{n} \frac{x_i x_i}{\sigma^2}}$$

As a result integrated likelihood has the following form:

$$p(y_i \mid x_i, \beta, \sigma^2) \propto \exp\{ \tfrac{1}{2} (-\sum_{i=1}^{n} \frac{x_i x_i}{\sigma^2} + (\beta_n / \tau_n^2 + \log(\tau_n^2)) - (\beta_0 / \tau_0^2 + \log(\tau_0^2)) \}$$

In fact for the case where we have already observed n data point, for a given state in HMM, the prior has the following form:

$$\beta \mid s_l \sim N(\beta_n, \tau_n^2)$$

$$\beta_n = \frac{\frac{\beta_0}{\tau_0^2} + r_{xy}^n}{\frac{1}{\tau_n^2} + r_{xx}^n}, r_{xy}^n = \sum_{i=1}^{n} \frac{y_i x_i s_{il}}{\sigma_l^2}$$

$$\tau_n^2 = \frac{1}{\frac{1}{\tau_0^2} + r_{xx}^n}, r_{xx}^n = \sum_{i=1}^{n} \frac{x_i x_i s_{il}}{\sigma_l^2}$$

where $s_{il}$ denotes the indicator variable for data point i that is equal to 1 when the data point is emitted from state $l$, and $\sigma_l^2$ denotes the variance of conditional distribution of emission from state $l$, and $r_{xy}^n$ and $r_{xx}^n$ are sufficient statistics of all n data points that have been observed so far.

As a result the posterior parameter after observing the n+1'th data point has the following form:

$$\beta \mid s_l \sim N(\beta_{n+1}, \tau_{n+1}^2)$$

$$\beta_{n+1} = \frac{\frac{\beta_0}{\tau_0^2} + r_{xy}^n + \frac{y_{n+1}x_{n+1}s_{n+1,l}}{\sigma_l^2}}{\frac{1}{\tau_0^2} + r_{xx}^n + \frac{x_{n+1}x_{n+1}s_{n+1,l}}{\sigma_l^2}}, r_{xy}^n = \sum_{i=1}^n \frac{y_i x_i s_l}{\sigma_l^2}$$

$$\tau_{n+1}^2 = \frac{1}{\frac{1}{\tau_0^2} + r_{xx}^n + \frac{x_{n+1}x_{n+1}s_{n+1,l}}{\sigma_l^2}}, r_{xx}^n = \sum_{i=1}^n \frac{x_i x_i s_l}{\sigma_l^2}$$

Consistent Rodriguez (2011), I integrate out the transition probability $\{\pi_{s_i}\}$, but to be able to run hierarchical DP model on the emission parameter, I do not integrate out the state-specific emission parameters $\{\Gamma_{s_i}\}$, so I draw particles for them along with other structural (i.e. non-state) parameters. Once the transition probabilities $\{\pi_{s_i}\}$ has been integrated out of the model, the transition distribution can be written as:

$$p(s_{it+1} \mid s_{it}, \dots, s_{i1}, \beta_i, \alpha_i) = \sum_{s_{it+1}=1}^{L_{it}} \frac{n_{s_{it}s_{it+1}}^t + \alpha_i \beta_{is_{it+1}}}{n_{s_{it}\cdot}^t + \alpha_i} \delta_{s_{it+1}} + \frac{\alpha_i \beta_{iL_{it+1}}}{n_{s_{it}\cdot}^t + \alpha_i} \delta_{L_{it+1}}$$

where

$L_{it} = \max\{s_{i1}, \dots, s_{it}\}$ denotes the number of distinct states visited by the process up to time t,

$n_{s_{it}s_{it+1}}^t$ denotes the number of transitions between state $s_{it}$ and $s_{it+1}$ up to time t, and

$n_{s_{it}\cdot}^t$ denotes the number of transitions out of state $s_{it}$ up to time t.

Conditional on the component of mixture normal that approximates logit model , the hidden motivation state of user $s_{it+1}$ at time t+1, and the component of hierarchical DP prior on the emission parameter $c_i$ the likelihood of observing data point $y_{it+1}$ weighted by the prior of emission parameter at time t+1 has the following form:

$$p(y_{it+1} \mid \Gamma_{s_{it}}^{pt}, \Lambda_i, c_i, s_{it+1}, U_{i|s_{it}}, z_{i|s_{it}}, r_{it}) = p_{Norm}(U_{i|s_{it}} \mid \Gamma_{s_{it}}, z_{i|s_{it}}, r_{it}, y_{it+1})$$

where $\Gamma_{s_{it}}^{pt}$ is a particle draw from posterior at time t.

As a consequence, the weighted one step ahead prediction distribution reduces to:

$$p(y_{it+1} \mid \Gamma_{s_{it}}, \Lambda_i, c_i, s_{it}, U_{i|s_{it}}, z_{i|s_{it}}, r_{it}) = \sum_{s_{it+1}=1}^{L_{it}} \frac{n_{s_{it}s_{it+1}}^{t} + \alpha_i \beta_{is_{it+1}}}{n_{s_{it}\cdot}^{t} + \alpha_i} \times p(y_{it+1} \mid \Gamma_{s_{it}}, \Lambda_i, c_i, s_{it+1}, U_{i|s_{it}}, z_{i|s_{it}}, r_{it})$$
$$+ \frac{\alpha_i \beta_{iL_{it+1}}}{n_{s_{it}\cdot}^{t} + \alpha_i} \times \left( p(y_{it+1} \mid \Gamma_{s_{it}}, \Lambda_i, c_i, s_{it+1}, U_{i|s_{it}}, z_{i|s_{it}}) \right)$$

Note that $s_{it+1}$ is the only unknown particle that we integrate over. As in our approach we have already taken particles for the emission parameter and its prior hyperparameters, integration over those priors are not required.

Note that sufficient statistics $r_{it}$ that we defined before is different for different hidden states, so formally it has the following structure:

$$\begin{aligned} r_{it} &= (r_{it}^{s_{i1}}, \ldots, r_{it}^{s_{iL_t}}) \\ r_{it}^{s_{ik}} &= (n_{s_{it}}, \overline{U}_{it}^{s_{ik}}, \overline{x}_{it}^{s_{ik}}, C_{U_{it}x_{it}}^{s_{ik}}, C_{x_{it}x_{it}}^{s_{ik}}, C_{U_{it}U_{it}}^{s_{ik}}) \\ n_{s_{it}} &= (n_{s_{it}1}, \ldots, n_{s_{it}s_{iL_t}}) \end{aligned}$$

Note that for the new state $L_{it} + 1$ no sufficient statistics exists, so only the prior parameters $\Delta D_i + \mu_{\Gamma i}$ and $\sigma_{\Gamma i}^2$ are relevant. Note that sufficient statistic vector $r_{it}$ is relevant only to compute posterior of structural parameters at time t, which becomes prior at time t+1, but I included it into this equation to emphasize that the information is already conditioned on by including the prior at time t+1.

Rodriguez (2011) suggests that consistent with MCMC algorithm series of auxiliary variables can be used to draw structural parameters. To draw $\beta = (\beta_1, \ldots, \beta_L, \beta_{L+1})$, a series of independent auxiliary variables $\{m_{ij}\}$ for states $i, j \in \{1,..,L\}$ can be sampled so that:

$$\Pr(m_{ij} = m \mid \ldots) \propto S(n_{ij}, m)(\alpha\beta_j)^m, m = 0, \ldots, n_{ij}$$

where $S(.,.)$ denotes the Stirling number of the first kind. Conditional on these auxiliary variables, $\beta$ can be sampled by:

$$\beta \mid \{m_{.i}\}, \lambda \sim Dir(m_{.1},...,m_{.L},\lambda)$$

where $m_{.j} = \sum_{i=1}^{L} m_{ij}$ .

Similarly, for shape parameter $\lambda$ , assuming $\lambda \sim Gam(a_\lambda, b_\lambda)$ a priori, another auxiliary variable $\phi$ can be introduced, so that $\phi \sim Beta(\lambda+1, m_{..})$. Under the gamma prior, the full conditional distribution for $\lambda$ given $\phi$ corresponds to a mixture of two gamma distributions as follows:

$$\lambda \mid a_\lambda, b_\lambda, L, \phi, \varepsilon \sim \varepsilon Gam(a_\lambda + L, b_\lambda - \log(\phi)) + (1-\varepsilon) Gam(a_\lambda + L - 1, b_\lambda - \log(\phi))$$

where $\frac{\varepsilon}{(1-\varepsilon)} = \frac{(a_\lambda + L - 1)}{\{m_{..}(b_\lambda - \log(\phi))\}}$.

In addition, for sampling the shape parameter $\alpha$ , another set of auxiliary variables can be used: $g_i \sim Beta(\alpha+1, n_{i.})$ and $h_i \sim Ber(\frac{n_{i.}}{\alpha + n_{i.}})$ for $i = 1,...,L$. Conditional on these latent variables, and $\alpha \sim Gam(a_\alpha, b_\alpha)$ a priori, the update of $\alpha$ has the following form:

$$\alpha \mid a_\alpha, b_\alpha, h_., m_{..}, g_{1..L} \sim Gam(a_\alpha + m_{..} - h_., b_\alpha - \sum_{i=1}^{L} \log g_i)$$

Note that we have also introduced two latent variables $U_{i|s_{it}}$ and $z_{i|s_{it}}$ to approximate logit emission probability model with mixture normal emission probability model. Frunwirth-Schnatter and Frunwirth-Schnatter (2007) also suggest using two extra auxiliary variables $d_i$ and $e_i$ from uniform distribution to sample latent variable $U_{i|s_{it}}$ as follows:

$$U_{i|s_{it}} = -\log\left( -\frac{\log d_i}{1+\exp(\Gamma_{s_{it}} x_{it})} - \frac{\log e_i}{\exp(\Gamma_{s_{it}} x_{it})} I_{\{y_{it}=0\}} \right)$$

Furthermore, to sample $z_{i|s_{it}}$ , the suggest the following structure:

$$\Pr(z_{i|s_{it}} = j \mid U_{i|s_{it}}, \Gamma_{s_{it}}) \propto \frac{w_j}{\sigma_{z_{i|s_{it}}}} \exp\left\{ -\frac{1}{2}\left( \frac{U_{i|s_{it}} - \Gamma_{s_{it}} x_{it} - \mu_{z_{i|s_{it}}}}{\sigma_{z_{i|s_{it}}}} \right)^2 \right\}$$

As these auxiliary and latent variables are required to draw other parameters, we can include them with the other structural parameters in a vector that is represented with particles, as follows:

$$\Theta_i^{(b)} = (\widehat{\alpha}_{it}^{(b)}, \{\widehat{\beta}_{lit}^{(b)}\}, \widehat{\lambda}_{it}^{(b)}, \{\widehat{\Gamma}_{lit}^{(b)}\}, \{\widehat{m}_{lit}^{(b)}\}, \widehat{\phi}_{it}^{(b)}, \{\widehat{h}_{lit}^{(b)}\}, \{\widehat{g}_{lit}^{(b)}\}, \widehat{U}_{i|s_{it}}^{(b)}, \widehat{z}_{i|s_{it}}^{(b)}, \widehat{d}_{i|s_{it}}^{(b)}, \widehat{e}_{i|s_{it}}^{(b)}, \widehat{\Lambda}_{i|s_{it}}^{(b)}, \widehat{c}_{i|s_{it}}^{(b)})$$

Additional particles that are required to be drawn are sufficient statistics for each state, and state indicator, as follows:

$$\Psi_i^{(b)} = (\widehat{s}_{it}^{(b)}, \widehat{L}_{it}^{(b)}, \{\widehat{r}_{lit}^{(b)}\})$$

Therefore, for the iHMM the particle learning algorithm for each user I has the following form:

**Algorithm1** Particle Learning Filtering

**(Initialize)**

Sample

$$\lambda_{i0}^{(b)} \sim Gam(a_\lambda, b_\lambda)$$
$$\alpha_{i0}^{(b)} \sim Gam(a_\alpha, b_\alpha)$$
$$\widehat{L}_{i0}^{(b)} = 0$$
$$\omega_0^{(b)} \propto 1$$

For all particles $b = 1..B$ set:

$$\widehat{L}_{i1}^{(b)} = 1$$
$$\widehat{s}_{i1}^{(b)} = 1$$
$$\widehat{n}_{111}^{(b)} = 1$$
$$r_{11}^{(b)} = r(y_{i1})$$
$$\hat{m}_{111}^{(b)} = 1$$

Sample $\widehat{\lambda}_{i1}^{(b)}$ by first sampling $\widehat{\phi}_{i1}^{(b)} \sim Beta(\lambda_{i0}^{(b)} + 1, \widehat{m}_{111}^{(b)})$ and then sampling $\widehat{\lambda}_{i1}^{(b)}$ from:

$$\widehat{\lambda}_{i1}^{(b)} \sim \varepsilon_1^{(b)} Gam(a_\lambda + \widehat{L}_{i1}^{(b)}, b_\gamma - \log(\widehat{\phi}_{i1}^{(b)})) + (1 - \varepsilon_1^{(b)}) Gam(a_\lambda + \widehat{L}_{i1}^{(b)} - 1, b_\gamma - \log(\widehat{\phi}_{i1}^{(b)}))$$

where

$$\frac{\varepsilon_1^{(b)}}{(1-\varepsilon_1^{(b)})} = \frac{a_\lambda + \widehat{L}_{i1}^{(b)} - 1}{\widehat{m}_{.,.,1}^{(b)}(b_\gamma - \log(\widehat{\phi}_{i1}^{(b)})}$$

Sample $\hat{\alpha}_{i1}^{(b)}$ by first generating for $l = \widehat{L}_{i1}^{(b)}$

$$\widehat{g}_{li1}^{(b)} \sim Beta(\hat{\alpha}_{i0}^{(b)} + 1, \widehat{n}_{i,l,.,1}^{(b)})$$

and

$$\widehat{h}_{i,l,1}^{(b)} \sim Ber(\frac{\widehat{n}_{i,l,.,1}^{(b)}}{\hat{\alpha}_{i0}^{(b)} + \widehat{n}_{i,l,.,1}^{(b)}})$$

and then

$$\widehat{\alpha}_{i1}^{(b)} \sim Gam(a_\alpha + \widehat{m}_{i,.,.,t+1}^{(b)} - \widehat{h}_{i,.,t+1}^{(b)}, b_\alpha - \sum_{l=1}^{\widehat{L}_{i1}^{(b)}} \log \widehat{g}_{i,l,1}^{(b)})$$

Sample

$$\widehat{\beta}_{i1}^{(b)} \sim Dir(\widehat{m}_{111}^{(b)}, \widehat{\lambda}_{i1}^{(b)})$$

Sample $c_{il1}^{(b)}$ for $l = \widehat{L}_{i1}^{(b)}, \widehat{L}_{i1}^{(b)} + 1$ from

$$c_{il1}^{(b)} \propto 1$$

Sample for $l = \widehat{L}_{i1}^{(b)}, \widehat{L}_{i1}^{(b)} + 1$

$$\widehat{\Lambda}_{il1}^{(b)} \sim N(\mu_\Delta D_i + \mu_{0B}, \Sigma_{0B})$$

Sample for $l = \widehat{L}_{i1}^{(b)}, \widehat{L}_{i1}^{(b)} + 1$

$$\Gamma_{il1}^{(b)} \sim N(\Delta_1^{(b)} D_i + \mu_{0B}, \Sigma_{0B})$$

Sample $U_{il}$ by first generating two auxiliary variables $d_{il1}$ and $e_{il1}$, for $l = \widehat{L}_{i1}^{(b)}$:

$$d_{il1}^{(b)} \sim Unif(0,1), e_{il1}^{(b)} \sim Unif(0,1),$$

and then

$$U_{il1}^{(b)} = -\log\left(-\frac{\log d_{il1}^{(b)}}{1+\exp(\Gamma_{il1}^{(b)} x_{i1})} - \frac{\log e_{il1}^{(b)}}{\exp(\Gamma_{il1}^{(b)} x_{i1})} I_{\{y_{i1}=0\}}\right)$$

Sample $z_{il}^{(b)}$ , for $l = \widehat{L}_{i1}^{(b)}$ by:

$$\Pr(z_{il1}^{(b)} = j \mid U_{il1}^{(b)}, \Gamma_{il1}^{(b)}) \propto \frac{w_j}{\sigma_{ilj1}^{(b)}} \exp\left\{-\frac{1}{2}\left(\frac{U_{il1}^{(b)} - \Gamma_{il1}^{(b)} x_{it} - \mu_{ilj1}^{(b)}}{\sigma_{ilj1}^{(b)}}\right)^2\right\}$$

**(Update)** : sufficient statistics for state $l = \widehat{s}_{i1}^{(b)}$ as:

$$\hat{\bar{x}}_i^{(1)} = x_{i1}$$
$$\hat{\bar{U}}_{i|s_{it}}^{(1)} = U_{i|s_{ii1}}^1$$
$$\widehat{C}_{ixx}^{(1)} = x_{i1}^2 - (\bar{x}_i^{(1)})^2$$
$$\widehat{C}_{xU_{i11}}^{(1)} = x_{i1}U_{i11}^1 - (\bar{x}_i^{(11)}\bar{U}_{i11}^{(1)})$$
$$\widehat{C}_{U_{i11}U_{i|s_{it}}}^{(1)} = (U_{i11}^1)^2 - (\bar{U}_{i11}^{(1)})^2$$

Draw the emission parameter $\Gamma_{il1}^{(b)}$ from its posterior:

$$\Gamma_{il1}^{(b)} \mid s_l \sim N(\mu_{\Gamma i1}^{(b)}, \sigma_{\Gamma i1}^{(b)^2})$$
$$\mu_{\Gamma i1}^{(b)} = \frac{\frac{\Lambda_{i,1,1}}{\exp(\Lambda_{i,2,1})} + \frac{r_{U_{il_t+1}^{(b)} x_{i11}}}{\sigma_{\tilde{z}_{il1}^{(b)}}^2}}{\frac{1}{\exp(\Lambda_{i,2,1})} + \frac{r_{U_{il_t+1}^{(b)} x_{i1}}}{\sigma_{\tilde{z}_{il1}^{(b)}}^2}}$$
$$\sigma_{\Gamma it}^{(b)^2} = \frac{1}{\frac{1}{\exp(\Lambda_{i,2,1})} + \frac{r_{U_{il_t+1}^{(b)} x_{i1}}}{\sigma_{\tilde{z}_{il1}^{(b)}}^2}}$$
$$\Lambda_{it} = (\mu_{\Gamma i1}, \log(\sigma_{\Gamma i1}^2))$$

**(main algorithm)**

**For** t=1 to T-1 **do**:

**(step-ahead prediction)** Compute weights $\omega^{(b)} = \frac{\vartheta_t^{(b)}}{\sum_{b'=1}^{B}\vartheta_t^{(b')}}$, where

$$\vartheta_{it}^{(b)} = \sum_{l=1}^{\widehat{L}_{it}^{(b)}+1} q_{il}^{(b)}(\widehat{\Psi}_{it}^{(b)}, \Theta_{it}^{(b)}, y_{it+1})$$
$$q_{il}^{(b)}(\widehat{\Psi}_{it}^{(b)}, \widehat{\Theta}_{it}^{(b)}, y_{it+1}) = \frac{\widehat{n}_{s_{it}s_{it+1}t}^{(b)} + \widehat{\alpha}_{it}^{(b)}\widehat{\beta}_{is_{it+1}t}^{(b)}}{\widehat{n}_{s_{it}.t}^{(b)} + \widehat{\alpha}_{it}^{(b)}} p(y_{it+1} \mid \widehat{\Gamma}_{s_{it+1}}^{(b)}, x_{it+1}, s_{it}^{(b)}, U_{s_{it+1}}^{(b)}, z_{s_{it+1}}^{(b)}), s_{it+1} \le \widehat{L}_{it}^{(b)}$$
$$q_{i,\widehat{L}_{it}^{(b)}+1}^{(b)}(\widehat{\Psi}_{it}^{(b)}, \widehat{\Theta}_{it}^{(b)}, y_{it+1}) = \frac{\widehat{\alpha}_{it}^{(b)}\widehat{\beta}_{iL_{it+1}}^{(b)}}{\widehat{n}_{s_{it}.t}^{(b)} + \widehat{\alpha}_{it}^{(b)}} \times p(y_{it+1} \mid \widehat{\Gamma}_{s_{it+1}}^{(b)}, x_{it+1}, s_{it}^{(b)}, U_{s_{it+1}}^{(b)}, z_{s_{it+1}}^{(b)}), s_{it+1} = \widehat{L}_{it}^{(b)} + 1$$

**(Re-sample)**: Sample $\left\{(\widetilde{\Psi}_{it}, \widetilde{\Theta}_{it})^{(b)}\right\}_{b=1}^{B}$ from current particles by generating an index $ind(b) \sim Multi(\omega^{(b)})$, so that

$$\widetilde{p}(\Psi_{it}, \Theta_{it} \mid y_{i1}, .., y_{it+1}) = \sum_{b=1}^{B}\omega^{(b)}\delta_{\widetilde{\Psi}_{it}^{(b)}, \widetilde{\Theta}_{it}^{(b)}}(\Psi_{it}, \Theta_{it})$$

**(Propagate)**: Propagate particles to generate $(\widehat{\Psi}_{it}^{(b)}, \widehat{\Theta}_{it}^{(b)})$ by:

(a) Sampling $\widehat{s}_{it+1}^{(b)}$ from $p(\widehat{s}_{it+1}^{(b)} \mid \widehat{s}_{it}^{(b)}, \widehat{\Theta}_{it}^{(b)}, y_{it+1})$, where

$$p(\widehat{s}_{it+1}^{(b)} \mid \widehat{s}_{it}^{(b)}, \widehat{\Theta}_{it}^{(b)}, y_{it+1}) \propto \sum_{l=1}^{\widetilde{L}_{it}^{(b)}+1} \frac{q_{il}^{(b)}(\widehat{\Psi}_{it}^{(b)}, \Theta_{it}^{(b)}, y_{it+1})}{\sum_{c=1}^{\widetilde{L}_{it}^{(b)}+1} q_{il}^{(c)}(\widehat{\Psi}_{it}^{(c)}, \Theta_{it}^{(c)}, y_{it+1})} \delta_{il}(\widehat{s}_{it+1}^{(b)})$$

(b) Update the number of states by setting

$$\widehat{L}_{it+1}^{(b)} = \begin{cases} \widetilde{L}_{it}^{(b)} + 1 & if \quad \widehat{s}_{it+1}^{(b)} = L_{it}^{(b)} + 1 \\ \widetilde{L}_{it}^{(b)} & if \quad otherwsie \end{cases}$$

(c) If $\widehat{s}_{it+1}^{(b)} \le \widetilde{L}_{it}^{(b)}$ set $\widehat{\beta}_{it+1}^{(b)} = \widetilde{\beta}_{it}^{(b)}$. Otherwise, update the transition probability vector by setting (STB)

$$\widehat{\beta}_{il,t+1}^{(b)} = \begin{cases} \widetilde{\beta}_{ilt}^{(b)} & if \quad l < \widetilde{L}_{it}^{(b)} \\ \Xi \widetilde{\beta}_{i\widetilde{L}_{it}^{(b)}+1,t}^{(b)} & if \quad l = \widetilde{L}_{it}^{(b)} \\ (1-\Xi)\widetilde{\beta}_{i\widetilde{L}_{it}^{(b)}+1,t}^{(b)} & if \quad l = \widetilde{L}_{it}^{(b)} + 1 \end{cases}$$

where $\Xi \sim Beta(1, \widetilde{\lambda}_{it}^{(b)})$

**(Update)** : sufficient statistics

(a) If $\widehat{s}_{it+1}^{(b)} \le \widetilde{L}_{it}^{(b)}$ update sufficient statistics for state $\widehat{s}_{it+1}^{(b)}$ as follows:

$$\hat{n}_{is_{it}s_{it+1},t+1}^{(b)} = \hat{n}_{is_{it}s_{it+1},t}^{(b)} + 1$$

otherwise create sufficient statistics for new state $\widehat{s}_{it+1}^{(b)} = \widetilde{L}_{it}^{(b)} + 1$ by

$$\hat{n}_{is_{it}s_{it+1},t+1}^{(b)} = 1$$

**(Sample)**: draw structural parameters and auxiliary variables(MCMC adaptation)

(a) Sample $\widehat{m}_{i,l,j,t+1}^{(b)} \in \{0,...,\hat{n}_{i,l,j,t+1}^{(b)}\}$ with

$$\Pr(\widehat{m}_{i,l,j,t+1}^{(b)} = m) \propto S(\hat{n}_{i,l,j,t+1}^{(b)}, m)(\widetilde{\alpha}_{it}^{(b)} \hat{\beta}_{i,j,t+1}^{(b)})^m$$

(b) Sample $\widehat{\lambda}_{it+1}^{(b)}$ by first sampling $\widehat{\phi}_{it+1}^{(b)} \sim Beta(\lambda_{it}^{(b)} + 1, \widehat{m}_{i,.,.,t+1}^{(b)})$ and then sampling $\widehat{\lambda}_{it+1}^{(b)}$ from:

$$\widehat{\lambda}_{it+1}^{(b)} \sim \varepsilon_{t+1}^{(b)} Gam(a_\lambda + \widehat{L}_{it+1}^{(b)}, b_\gamma - \log(\widehat{\phi}_{it+1}^{(b)})) + (1-\varepsilon_{t+1}^{(b)}) Gam(a_\lambda + \widehat{L}_{it+1}^{(b)} - 1, b_\gamma - \log(\widehat{\phi}_{it+1}^{(b)}))$$

where

$$\frac{\varepsilon_{t+1}^{(b)}}{(1-\varepsilon_{t+1}^{(b)})} = \frac{a_\lambda + \widehat{L}_{it+1}^{(b)} - 1}{\widehat{m}_{i,.,.,t+1}^{(b)}(b_\gamma - \log(\widehat{\phi}_{it+1}^{(b)})}$$

Sample $\hat{\alpha}_{it+1}^{(b)}$ by first generating for $l = 1,..,\widehat{L}_{it+1}^{(b)}$

$$\widehat{g}_{lit+1}^{(b)} \sim Beta(\hat{\alpha}_{it}^{(b)} + 1, \widehat{n}_{i,l,.,t+1}^{(b)})$$

and

$$\widehat{h}_{ilt+1}^{(b)} \sim Ber(\frac{\widehat{n}_{i,l,.,t+1}^{(b)}}{\hat{\alpha}_{i0}^{(b)} + \widehat{n}_{i,l,.,t+1}^{(b)}})$$

and then

$$\widehat{\alpha}_{it+1}^{(b)} \sim Gam(a_\alpha + \widehat{m}_{i,.,.,t+1}^{(b)} - \widehat{h}_{i,.,t+1}^{(b)}, b_\alpha - \sum_{l=1}^{\widehat{L}_{it+1}^{(b)}} \log \widehat{g}_{i,l,t+1}^{(b)})$$

Re-sample

$$\widehat{\beta}_{it+1}^{(b)} \sim Dir(\widehat{m}_{i,.,1,t+1}^{(b)},..,\widehat{m}_{i,.,\widehat{L}(b)_{it+1},t+1}^{(b)}, \widehat{\lambda}_{it+1}^{(b)})$$

**(Sample)**: sample structural emission parameters given their posterior distributions, conditioned on information available up to time t for state $\widehat{s}_{it+1}^{(b)}$

(a) Sample $U_{ilt+1}$ by first generating two auxiliary variables $d_{ilt+1}$ and $e_{ilt+1}$, for state $l = \widehat{s}_{it+1}^{(b)}$:

$$d_{ilt+1}^{(b)} \sim Unif(0,1), e_{ilt+1}^{(b)} \sim Unif(0,1),$$

and then

$$U_{ilt+1}^{(b)} = -\log\left(-\frac{\log d_{ilt+1}^{(b)}}{1+\exp(\Gamma_{ilt}^{(b)} x_{it+1})} - \frac{\log e_{ilt+1}^{(b)}}{\exp(\Gamma_{ilt}^{(b)} x_{it+1})} I_{\{y_{it+1}=0\}}\right)$$

(b) Draw the component index for the latent utility $\widehat{z}_{il\,t+1}^{(b)}$ for state $l = \widehat{s}_{it+1}^{(b)}$ as

$$\widehat{z}_{il\,t+1}^{(b)} = z \mid U_{il\,t+1}^{(b)}, \Gamma_{il\,t}^{(b)}, y_{t+1} \sim Dir(w_1 p_N(z=1, y_{it+1}, U_{il\,t+1}^{(b)}, \mu_{z=1}, \sigma_{z=1}, x_{it+1}, \Gamma_{il\,t}^{(b)}),..,$$
$$w_{10} p_N(z=10, y_{it+1}, U_{il\,t+1}^{(b)}, \mu_{z=10}, \sigma_{z=10}, x_{it+1}, \Gamma_{il\,t}^{(b)}))$$

or simply:

$$\Pr(z_{il1}^{(b)} = j \mid U_{il\,t+1}^{(b)}, \Gamma_{ilt}^{(b)}) \propto \frac{w_j}{\sigma_{iljt}^{(b)}} \exp\left\{-\frac{1}{2}\left(\frac{U_{il\,t+1}^{(b)} - \Gamma_{ilt}^{(b)} x_{it+1} - \mu_{iljt}^{(b)}}{\sigma_{iljt}^{(b)}}\right)^2\right\}$$

(c) **(Update)** : sufficient statistics for state $l = \widehat{s}_{it+1}^{(b)}$ as:

If $\widehat{s}_{it+1}^{(b)} \le \widetilde{L}_{it}^{(b)}$ update sufficient statistics for state $\widehat{s}_{it+1}^{(b)}$ as follows:

$$\hat{\overline{x}}_i^{(t+1)} = \overline{x}_i^{(t)} + \frac{1}{t+1}(x_{it+1} - \overline{x}_i^{(t)})$$

$$\widehat{\overline{U}}_{i|s_{it}}^{(t+1)} = \overline{U}_{i|s_{it}}^{(t)} + \frac{1}{t+1}(U_{i|s_{it+1}}^{t+1} - \overline{U}_{i|s_{it}}^{(t)})$$

$$\widehat{C}_{ixx}^{(t+1)} = \frac{t}{t+1}(C_{ixx}^{(t)} + (\overline{x}_i^{(t)})^2) + (\frac{x_{it+1}^2}{t+1}) - (\overline{x}_i^{(t+1)})^2$$

$$\widehat{C}_{xU_{i|s_{it}}}^{t+1} = \frac{t}{t+1}(C_{xU_{i|s_{it}}}^{(t)} + (\overline{x}_i^{(t)}\overline{U}_{i|s_{it}}^{(t)})) + (\frac{x_{it+1}U_{i|s_{it+1}}^{t+1}}{t+1}) - (\overline{x}_i^{(t+1)}\overline{U}_{i|s_{it}}^{(t+1)})$$

$$\widehat{C}_{U_{i|s_{it}}U_{i|s_{it}}}^{(t+1)} = \frac{t}{t+1}(C_{U_{i|s_{it}}U_{i|s_{it}}}^{(t)} + (\overline{U}_{i|s_{it}}^{(t)})^2) + \frac{(U_{i|s_{it+1}}^{t+1})^2}{t+1} - (\overline{U}_{i|s_{it}}^{(t+1)})^2$$

otherwise create sufficient statistics for new state $\widehat{s}_{it+1}^{(b)} = \widetilde{L}_{it}^{(b)} + 1$ by

$$\hat{\overline{x}}_i^{(t+1)} = x_{it+1}$$

$$\widehat{\overline{U}}_{i|s_{it}}^{(t+1)} = U_{i|s_{it+1}}^{t+1}$$

$$\widehat{C}_{ixx}^{(t+1)} = x_{it+1}^2 - (\overline{x}_i^{(t+1)})^2$$

$$\widehat{C}_{xU_{i|s_{it}}}^{t+1} = x_{it+1}U_{i|s_{it+1}}^{t+1} - (\overline{x}_i^{(t+1)}\overline{U}_{i|s_{it}}^{(t+1)})$$

$$\widehat{C}_{U_{i|s_{it}}U_{i|s_{it}}}^{(t+1)} = (U_{i|s_{it+1}}^{t+1})^2 - (\overline{U}_{i|s_{it}}^{(t+1)})^2$$

(d) Draw the emission parameter $\Gamma_{il\,t+1}^{(b)}$ from its posterior for $l = \widehat{s}_{it+1}^{(b)}$:

$$\Gamma_{il\,t+1}^{(b)} \mid s_l \sim N(\mu_{\Gamma it}^{(b)}, \sigma_{\Gamma it}^{(b)^2})$$

$$\mu_{\Gamma it}^{(b)} = \frac{\dfrac{\Lambda_{i,1,t}}{\exp(\Lambda_{i,2,t})} + \dfrac{r_{U_{il_{t+1}}^{(b)}x_{it+1}}}{\sigma_{\tilde{z}_{il_{t+1}}^{(b)}}^2}}{\dfrac{1}{\exp(\Lambda_{i,2,t})} + \dfrac{r_{U_{il_{t+1}}^{(b)}x_{it+1}}}{\sigma_{\tilde{z}_{il_{t+1}}^{(b)}}^2}}$$

$$\sigma_{\Gamma it}^{(b)^2} = \frac{1}{\dfrac{1}{\exp(\Lambda_{i,2,t})} + \dfrac{r_{U_{il_{t+1}}^{(b)}x_{it+1}}}{\sigma_{\tilde{z}_{il_{t+1}}^{(b)}}^2}}$$

**(Hierarchical VB)**

(a) Draw $\Delta^{(b)}$ same across all users from its posterior:

$$\Delta_{t+1}^{(b)} \sim N(\mu_{\Delta,t+1}, \Sigma_{\Delta,t+1})$$

$$\mu_{\Delta,t+1} = \frac{\frac{\mu_{\Delta,t}}{\Sigma_{\Delta,t}} + \sum \frac{\widehat{\Lambda}_{ilt+1}^{(b)} D_i}{\sigma_{c_{ilt}}^{(b)2}}}{\frac{1}{\Sigma_{\Delta,t}} + \frac{D_i D_i}{\sigma_{c_{ilt}}^{(b)2}}}$$

$$\Sigma_{\Delta,t+1} = \frac{1}{\frac{1}{\Sigma_{\Delta,t}} + \frac{D_i D_i}{\sigma_{c_{ilt}}^{(b)2}}}$$

(b) Run Variational Bayesian for DP across all users given $\left\{ \frac{\sum_{b=1}^{B} \widehat{\Lambda}_{ilt}^{(b)} - \Delta_{t+1}^{(b)} D_i}{B} \right\}$, and recover $\{\widehat{q}(\eta_{kt} \mid \tau_{\eta kt}), \widehat{q}(c_{it} \mid \phi_{cit}), \widehat{q}(v_{kt} \mid \gamma_{vkt})\}$

(a) Draw prior for the emission parameter distribution based on variational posterior as follows

Sample $c_{ilt+1}^{(b)}$ for $l = 1 ... \widehat{L}_{it+1}^{(b)}, \widehat{L}_{it+1}^{(b)} + 1$ from

$$c_{ilt+1}^{(b)} \propto \widehat{q}(c_{ilt} \mid \phi_{it})$$

Sample for $l = 1 ... \widehat{L}_{it+1}^{(b)}, \widehat{L}_{it+1}^{(b)} + 1$

$$\widehat{\Lambda}_{ilt+1}^{(b)} \sim N(\Delta_{t+1}^{(b)} D_i + \mu_{\Lambda c_{ilt+1}^{(b)}}, \Sigma_{\Lambda c_{ilt+1}^{(b)}})$$

(b) Draw the emission parameter $\Gamma_{il\,t+1}^{(b)}$ from its prior for $l = \widetilde{L}_{it}^{(b)} + 1$:

$$\Gamma_{ilt+1}^{(b)} \sim N(\widehat{\Lambda}_{ilt+1,1}^{(b)}, \widehat{\Lambda}_{ilt+1,2}^{(b)})$$

**End for**

Finally, it is relevant to note that the above PL except the Hierarchical DP VB can be run in parallel to speed up the estimation procedure. Finally, we initialize the procedure with uninformative/ vague prior to get reliable estimates. This procedure gives a time evolving posterior which is approximation to the true/target posterior. Each posterior is updated in the light of recent observations. About the identification we have to note that per exchangeability property the states are subject to label switching.

**Variational Bayesian for Dirichlet Process Prior**

As Blei and Jordan (2006) suggest, the DP can be used for nonparametric prior in a hierarchical Bayesian model. The process in a general form looks as follows:

$$
\begin{aligned}
&G \sim DP(\alpha, G_0) \\
&\eta_n \sim G \\
&X_n \sim p(.\,|\,\eta_n)
\end{aligned}
$$

where $\alpha$ is scaling parameter, and G0 is baseline Dirichlet distribution. As the parameter are drawn from G, the data themselves will partition according to the drawn values from the same parameters. It is a form of infinite mixture model, in which we draw the parameters either from one of the partitions of parameters we have seen before, or from a new partition. This process is sometimes referred to as Polya's urn or Chinese restaurant process. Another view suggests a stick breaking construction of G, by considering $V_i \sim Beta(1,\alpha)$ and $\eta_i^* \sim G$ for $i = \{1,2,...\}$. As a result formally we can define G and the proportions $\theta_i$ of each of the infinite pieces of stick relative to original unit-length stick with size proportional to number of draws from a distribution as:

$$
\begin{aligned}
&\theta_i = V_i \prod\nolimits_{j=1}^{i-1} (1 - V_i) \\
&G(\eta) = \sum\nolimits_{i=1}^{\infty} \theta_i \delta(\eta, \eta_i^*)
\end{aligned}
$$

As Blei and Jordan (2006) suggest, $\theta$ comprises the infinite vector of mixing proportions and $\eta_{1:\infty}^*$ are the infinite number of mixture components. We denote $Z_n$ as the mixture component with which $X_n$ is associated. Therefore the data generating process for DP is as follows:

1. Draw $V_i \sim Beta(1,\alpha), i = \{1,2,...\}$

2. Draw $\eta_i \sim G_0,, i = \{1,2,...\}$

3. For each data point n:

   a. Draw $Z_n \sim Mult(\theta)$

   b. Draw $X_n \sim F(\eta_{z_n})$

To estimate this model Blei and Jordan (2006) suggest we truncate this construction at K, by setting $V_{K-1} = 1$ , which translates into $\theta_k = 0, k > K$. It has shown that the truncated Dirichlet process (TDP), closely approximates a true Dirichlet process for K chosen large enough relative to the number of data. To estimates this model we use Variational Bayesian (VB) approximate of variational distribution, and its parameters. VB uses optimization of variational distribution (with free parameters) rather than sampling like Gibbs sampler. As we have selected the blocks of our model conjugate to each other, we can use mean field VB (MFVB) rather than fixed form VB (FFVB), which is appropriate for non-conjugate models. MFVB is closely related to Gibbs sampling, but it does not have the problem of mixing and stickiness that Gibbs sampler has.

The Jensen's inequality suggests, a lower bound for log-likelihood as:

$$\log p(x) = \log \int_h p(x,h)dh = \log \int_h \frac{q(h)p(x,h)}{q(h)} dh \geq \int_h q(h)\log(p(x,h))dh - \int_h q(h)\log(q(h))dh$$
$$= E_q[\log p(x,H)] - E_q[\log q(H)]$$

The above inequality can intuitively be explained by the concavity of the log function, and it should be satisfied with an arbitrary distribution q(h). H denotes the hidden variables (including unknown parameters), and x denotes the observations. The $E_q[\log p(x,H)] - E_q[\log q(H)]$ is called the evidence lower bound (ELBO). The first element of ELBO $E_q[\log p(x,H)]$ is called the energy function, and the second one is the entropy of the variational distribution $E_q[\log q(H)]$. Formally, the relationship between K-L divergence and the evidence lower bound can be demonstrated as follows:

$$\begin{aligned} KL[q(H) \| p(H \mid x)] &= E_q[\log q(H)] - E_q[\log p(H \mid x)] \\ &= -E_q[\log p(x, H)] + E_q[\log q(H)] + \log p(x) \end{aligned}$$

The key trick behind variational methods is to restrict q(h) to a parametric family such that optimizing the bound is tractable. The solution is usually straight forward by considering the natural parameter and sufficient statistic of specific family of distributions. Penny (2001) computes Kullback-Leibler (KL) divergence or relative entropy of of Normal, Gamma, Dirichlet and Wishart densities. For DP mixture Blei and Jordan (2006) apply mean filed variational approach for the stick-breaking construction. Hidden variables and unknown parameters of the model are $V$ (stick breaking construction parameter that builds mixing distribution), $\eta^*$ (the prior on the distribution of mean and variance of each partition), and Z (the index of partition membership for each observation), and coupling them in the likelihood makes it analytically intractable. Thus, we have to introduce a variational distribution $q(v, \eta^*, z)$, in which all the hidden variables are independent, as we factorize this variational distribution. As a result our factorized variational distribution can be written as:

$$q(v, \eta^*, z, K) = \prod_{i=1}^{K} q(v_i \mid \gamma_i) \prod_{i=1}^{K} q(\eta_i^* \mid \tau_i) \prod_{n=1}^{N} q(z_n \mid \phi_n)$$

where $\gamma$ are the Beta parameters for the distributions on $V_i$ (stick breaking construction parameter that builds mixing distribution), $\tau$ are natural parameters for the distributions on $\eta_i^*$ (the prior on the distribution of mean and variance of each partition), and $\phi$ are multinomial parameters for the distribution on $Z_n$ (the index of partition membership for each observation). Therefore the lower bound on the likelihood by K-L divergence criteria can be written as:

$$\log p(x \mid \alpha, \lambda) \geq E[\log p(V \mid \alpha)] + E[\log p(\eta^* \mid \lambda)] + \sum_{n=1}^{N} E[\log p(Z_n \mid V)]$$
$$+ E[\log(p(x_n \mid Z_n)] - E[\log q(Z, V, \eta^*)]$$
$$E[\log p(Z_n \mid V)] = E[\log(\prod_{i=1}^{K} (1 - V_i)^{1[Z_n > i]} V_i^{Z_n^i}$$

We need the following elements to compute the K-L divergence:

$$E[\log p(Z_n \mid V)] = \sum_{i=1}^{k} q(z_n > i) E[\log(1 - V_i] + q(z_n = i) E[\log V_i]$$
$$q(z_n = i) = \phi_{n,i}$$
$$q(z_n > i) = \sum_{j=i+1}^{K} \phi_{n,i}$$
$$E[\log V_i] = \Psi(\gamma_{i,1}) - \Psi(\gamma_{i,1} + \gamma_{i,2})$$
$$E[\log(1 - V_i)] = \Psi(\gamma_{i,2}) - \Psi(\gamma_{i,1} + \gamma_{i,2})$$

Optimization of K-L divergence criteria can be done by a coordinate ascent algorithm in the variational parameters. Coordinate ascent for exponential family distributions iteratively sets each natural variational parameter equal to the expectation of the natural conditional parameter given all other variables and observations. The algorithm is derived by equating the first order condition of the K-L divergence (or its corresponding evidence lower bound) with respect to the variational distribution to zero (by including the Largrangian multiplier condition that the variational distribution shall integrate to one). Formally:

$$\frac{\partial ELBO}{\partial q} = 0$$
$$q_i^*(H_i) \propto \exp\{E_{H_{-i}}[\log p(H_i \mid x, H_{-i})]\}$$

The updates of $\gamma_n$ follow the standard recipe for variational inference with exponential family distribution in a conjugate setting (Ghahramani & Beal, 2001), so for the parameters of the beta distribution $V_i$ (stick breaking construction parameter that builds mixing distribution), we have:

$$\gamma_{i,1} = 1 + \sum_{m=1}^{N} \phi_{n,i}$$
$$\gamma_{i,2} = \alpha + \sum_{k=i+1}^{K} \sum_{m=1}^{N} \phi_{n,i}$$

The update for the variational multinomial parameter $\phi_{n,i}$ of the distribution of the membership index for each observation $Z_n$, with parameter is proportional to:

$$\exp(E[\log V_i \mid \gamma_i] + E[\eta_i \mid \tau_i]^T X_n - E[a(\eta_i) \mid \tau_i] - \sum_{j=i+1}^{K} E[\log(1 - V_i) \mid \gamma_j])$$

For the Gaussian component portion, we adopted an algorithm suggested by Penny (2002). We refer interested reader to that short instruction. For the model of this paper, I start with defining the prior on the parameters as follows (note that index is time varying as the variational Bayesian procedure is run at each point in time when new information becomes available as a result of running particle learning algorithm):

The prior on the mixing distribution, which has tick breaking construction, is defined as:

$$\alpha_{v0} \sim Gamma(a_{v0}, b_{v0})$$
$$v_{k0} \sim Beta(1, \alpha_{v0})$$
$$\theta_{k0} \sim STB(V_0) \equiv \theta_{k0} = v_{k0} \prod_{j=1}^{k-1} (1 - v_{j0})$$
$$\alpha_{vt}^{-1} \sim Gamma(a_{vt}^{-1}, b_{vt}^{-1})$$
$$v_{kt}^{-1} \sim Beta(\gamma_{vkt1}^{-1}, \gamma_{vkt2}^{-1})$$
$$\theta_{kt}^{-1} \sim STB(V_t^{-1}) \equiv \theta_{kt}^{-1} = v_{kt}^{-1} \prod_{j=1}^{k-1} (1 - v_{jt}^{-1})$$

The prior on the distribution of precision parameter of each partition, which has Wishart distribution, is defined as:

$$p(\Sigma_{\Lambda k0}) = W(a_{\Lambda\Sigma 0}, B_{\Lambda\Sigma 0})$$
$$p(\Sigma_{\Lambda kt}^{-1}) = W(a_{\Lambda\Sigma kt}^{-1}, B_{\Lambda\Sigma kt}^{-1})$$

The prior over the mean parameter of the partitions given the precision parameter, which has normal distribution, is defined as:

$$p(\mu_{\Lambda k0} \mid \Sigma_{\Lambda k0}) = N(m_{\Lambda\mu 0}, \beta_{\Lambda\mu 0}\Sigma_{\Lambda k0})$$
$$p(\mu_{\Lambda kt}^{-1} \mid \Sigma_{\Lambda kt}^{-1}) = N(m_{\Lambda\mu kt}^{-1}, \beta_{\Lambda\mu kt}^{-1}\Sigma_{\Lambda kt}^{-1})$$

The joint likelihood of data points and partition membership indicator has the following form:

$$p(\Lambda_{ilt}, c_{ilt}) = p(c_{ilt} = k \mid \theta_{kt}^{-1}) p(\Lambda_{ilt} \mid \mu_{\Lambda kt}^{-1}, \Sigma_{\Lambda kt}^{-1})$$

The variational approximation to the posterior of the mixing distribution, which has tick breaking construction, is defined as:

$$q(\alpha_{vt}) = Gamma(a_{vt}, b_{vt})$$
$$q(v_{kt}) = Beta(\gamma_{vkt1}, \gamma_{vkt2})$$
$$\theta_{kt} \sim STB(V_t) \equiv \theta_{kt} = v_{kt}\prod_{j=1}^{k-1}(1 - v_{jt})$$

where

$a_{vt}$ and $b_{vt}$ denote the variational parameters of gamma posterior distribution of concentration parameter.

$\gamma_{vkt1}$ and $\gamma_{vkt2}$ denote the variational parameters of the beta posterior distribution of the parameter of stick breaking construction of mixture proportion distribution.

The variational approximation to the posterior of partition precisions has the following form:

$$q(\Sigma_{\Lambda kt}) = W(a_{\Lambda\Sigma kt}, B_{\Lambda\Sigma kt})$$

Where

$a_{\Lambda\Sigma kt}$ and $B_{\Lambda\Sigma kt}$ denote the variational parameter of Wishart posterior distribution of the precision parameters of the partitions.

The variational approximation to the posterior of the partition means given precisions has the following form:

$$q(\mu_{\Lambda kt} \mid \Sigma_{\Lambda kt}) = N(m_{\Lambda\mu kt}, \beta_{\Lambda\mu kt}\Sigma_{\Lambda kt})$$

where

$m_{\Lambda\mu kt}$ and $\beta_{\Lambda\mu t}$ denote the variational parameter of multivariate normal posterior distribution of the mean parameters of the partitions.

The variational approximation to the posterior of the partition membership indicator for each emission parameter (or assignment probability) of user i at state l has the following form:

$$q(c_{ilt} \mid \phi_{ct}) \sim Mult(\phi_{ci1t}, \ldots, \phi_{ciKt})$$

where

$\phi_{ci1t}, \ldots, \phi_{ciKt}$ denote the variational parameter of multinomial posterior distribution of the partition membership index.

The coordinate ascent algorithm to estimate variational distribution as a proxy for the posterior of the parameters follows the iterations of Expectation Maximization (EM) algorithm structure as follows:

**Algorithm3** Variational Expectation Maximization (VEM)

**(Initialization-step)**

To allow the maximum possible partitions (segments) set the number of partitions to the number of data points (emission parameters) $K = I * L$

Create random uniform number for partition membership of each data point (emission parameter) $\tilde{\phi}_{cilkt} \sim Unif(0,1)$

Create the probability of partitions membership probability for each data point by normalizing $\tilde{\phi}_{cilkt}$ as follows:

$$\phi_{cilkt} = \frac{\tilde{\phi}_{cilkt}}{\sum_{k'=1}^{K} \tilde{\phi}_{cilk't}}$$

Set the concentration parameter

$$\alpha_{vt} = \frac{a_{vt}^{-1}}{b_{vt}^{-1}}$$

Set the hyper-parameters of the stick breaking construction by

$$\gamma_{vkt1} = \gamma_{vkt1}^{-1}$$
$$\gamma_{vkt2} = \alpha_{vt}$$

Set the hyper-parameters of the mean of the partitions by

$$\beta_{\Lambda\mu kt} = \beta_{\Lambda\mu kt}^{-1}$$

$$m_{\Lambda\mu kt} = m_{\Lambda\mu kt}^{-1}$$

Set the hyper-parameters of the precision of the partitions by

$$B_{\Lambda\Sigma kt} = B_{\Lambda\Sigma kt}^{-1}$$
$$a_{\Lambda\mu kt} = a_{\Lambda\mu kt}^{-1}$$

**(E-step)** Update the posterior for the partition (segment) membership indicator multinomial distribution ( the probability that k'th partition is responsible for i'th emission parameter in l'th state) by

$$\phi_{cilkt} = \frac{\widetilde{\phi}_{cilkt}}{\sum_{k'=1}^{K} \widetilde{\phi}_{cilk't}}$$

where for computation of probability until the normalization constant we have

$$\widetilde{\phi}_{cilkt} = \widetilde{\theta}_{kt} \widetilde{\Sigma}_{\Lambda kt}^{1/2} \exp\left[ -\frac{1}{2} (\Lambda_{ilt} - m_{\Lambda \mu kt})^T \overline{\Sigma}_{\Lambda kt} (\Lambda_{ilt} - m_{\Lambda \mu kt}) \right] \exp\left[ \frac{-D}{2\beta_{\Lambda \mu kt}} \right]$$

$$D = \dim(\Lambda_{ilt})$$

$$\log(\widetilde{\theta}_{kt}) = E[\log \nu_k \mid \gamma_{vkt}] - \sum_{j=k+1}^{K} E[\log(1-\nu_k) \mid \gamma_{vkt}] = \Psi_\gamma(\gamma_{vkt1}) - \Psi_\gamma(\gamma_{vkt1} + \gamma_{vkt2}) - \sum_{j=k+1}^{K} \Psi_\gamma(\gamma_{vjt2}) - \Psi_\gamma(\gamma_{vjt1} + \gamma_{vjt2})$$

$$\log \widetilde{\Sigma}_{\Lambda kt} = \sum_{d=1}^{D} \Psi_\gamma((a_{\Lambda \Sigma kt} + 1 - d)/2) - \log |B_{\Lambda \Sigma kt}| + D \log 2$$

$$\overline{\Sigma}_{\Lambda kt} = a_{\Lambda \Sigma kt} B_{\Lambda \Sigma kt}^{-1}$$

and $\Psi_\gamma(.)$ denotes the digamma function.

Update posterior for the concentration parameter of DP by

$$a_{vt} = a_{vt}^{-1} + K - 1$$

$$b_{vt} = b_{vt}^{-1} - \sum_{k=1}^{K-1} \Psi_\gamma(\gamma_{vkt2}) - \Psi_\gamma(\gamma_{vkt1} + \gamma_{vkt2})$$

**(M-step)**

First, we define the following:

$$\overline{\theta}_{kt} = \frac{1}{I * L} \sum_{i,l=1}^{I*L} \phi_{cilkt}$$

$$\overline{N}_{kt} = \sum_{i,l=1}^{I*L} \phi_{cilkt}$$

$$\overline{\mu}_{\Lambda kt} = \frac{1}{\overline{N}_{kt}} \sum_{i,l=1}^{I*L} \phi_{cilkt} \Lambda_{ilt}$$

$$\overline{\Sigma}_{\Lambda kt} = \frac{1}{\overline{N}_{kt}} \sum_{i,l=1}^{I*L} \phi_{cilkt} (\Lambda_{ilt} - m_{\Lambda \mu kt})(\Lambda_{ilt} - m_{\Lambda \mu kt})^T$$

where

$\overline{\theta}_{kt}$ denotes the proportion of data points (emission parameters) in partition k at time t

$\overline{N}_{kt}$ denotes the number of data points (emission parameters) in partition k at time t

$\overline{\mu}_{\Lambda kt}, \overline{\Sigma}_{\Lambda kt}$ denote mean and variance of data points (emission parameters) in partition k at time t

We update the hyperparameters as follows:

Update hyperparameters of the variational distribution of the Stick Breaking Construction as

follows:

$$\gamma_{vkt1} = \gamma_{vkt1}^{-1} + \sum_{i,l=1}^{I,L} \phi_{cilkt}$$

$$\gamma_{vkt2} = \frac{a_{vt}}{b_{vt}} + \sum_{i,l=1}^{I,L} \sum_{j=k+1}^{K} \phi_{ciljt}$$

Update hyperparameters of the variational distribution of mean of partitions as follows:

$$m_{\Lambda\mu kt} = \frac{\beta_{\Lambda\mu kt}^{-1} m_{\Lambda\mu kt}^{-1} + \overline{N}_{kt} \overline{\mu}_{\Lambda kt}}{\overline{N}_{kt} + \beta_{\Lambda\mu kt}^{-1}}$$

$$\beta_{\Lambda\mu kt} = \overline{N}_{kt} + \beta_{\Lambda\mu kt}^{-1}$$

Update the hyperparameters of variational distribution of precision of partitions as follows:

$$B_{\Lambda\Sigma kt} = B_{\Lambda\Sigma kt}^{-1} + \frac{\overline{N}_{kt} \beta_{\Lambda\mu kt}^{-1} (\overline{\mu}_{\Lambda kt} - m_{\Lambda\mu kt}^{-1})(\overline{\mu}_{\Lambda kt} - m_{\Lambda\mu kt}^{-1})^T}{\overline{N}_{kt} + \beta_{\Lambda\mu kt}^{-1}} + \overline{N}_{kt} \overline{\Sigma}_{\Lambda kt}$$

$$a_{\Lambda\mu kt} = \overline{N}_{kt} + a_{\Lambda\mu kt}^{-1}$$

**(iterations)**

Iterated between E-step and M-step until Evidence Lower Bound (ELBO) is maximized (converged), or equivalently the K-L divergence is minimized, as follows:

$$ELBO = \int q(\alpha_{vt}) \log \frac{p(\alpha_{vt})}{q(\alpha_{vt})} d\alpha_{vt} + \sum_{k=1}^{K} \int q(\nu_{kt}) \log \frac{p(\nu_{kt})}{q(\nu_{kt})} d\nu_{kt} + \sum_{k=1}^{K} \int q(\Sigma_{\Lambda kt}) \log \frac{p(\Sigma_{\Lambda kt})}{q(\Sigma_{\Lambda kt})} d\Sigma_{\Lambda kt} +$$
$$\sum_{k=1}^{K} \int q(\mu_{\Lambda kt} | \Sigma_{\Lambda kt}) \log \frac{p(\mu_{\Lambda kt} | \Sigma_{\Lambda kt})}{q(\mu_{\Lambda kt} | \Sigma_{\Lambda kt})} d\mu_{\Lambda kt} + \sum_{k=1}^{K} \sum_{i,l=1}^{I,L} q(c_{ilt}) \int q(\nu_{kt}) \log \frac{p(c_{ilt} | \nu_{kt})}{q(c_{ilt})} d\nu_{kt} +$$
$$\sum_{k=1}^{K} \sum_{i,l=1}^{I,L} q(c_{ilt}) \int\int q(\Sigma_{\Lambda kt}) q(\mu_{\Lambda kt} | \Sigma_{\Lambda kt}) \log \frac{p(\Lambda_{ilt} | \Sigma_{\Lambda kt}, \mu_{\Lambda kt}, c_{ilt})}{q(c_{ilt})} d\Sigma_{\Lambda kt} d\mu_{\Lambda kt}$$

Writing the ELBO in terms of K-L divergence leads to:

$$ELBO = -KL_{Gamma}(a_{vt}, b_{vt}; a_{vt}^{-1}, b_{vt}^{-1}) - \sum_{k=1}^{K} KL_{Beta}(\gamma_{vkt1}, \gamma_{vkt2}; \gamma_{vkt1}^{-1}, \gamma_{vkt2}^{-1}) - \sum_{k=1}^{K} KL_{W}(a_{\Lambda\Sigma kt}, B_{\Lambda\Sigma kt}; a_{\Lambda\Sigma kt}^{-1}, B_{\Lambda\Sigma kt}^{-1})$$
$$- \sum_{k=1}^{K} KL_{N}(m_{\Lambda\mu kt}, \frac{\Sigma_{\Lambda kt}}{\beta_{\Lambda\mu kt} a_{\Lambda\Sigma kt}}; m_{\Lambda\mu kt}^{-1}, \frac{\Sigma_{\Lambda\mu kt}^{-1}}{\beta_{\Lambda\mu kt}^{-1} a_{\Lambda\mu kt}^{-1}})$$
$$+ \sum_{k=1}^{K} \overline{N}_{kt} \log(\widetilde{\theta}_{kt}) - \sum_{i,l=1}^{I,L} \phi_{cilkt} \log(\phi_{cilkt}) + \frac{\overline{N}_{kt}}{2}(-D \log 2\pi + \log \widetilde{\Sigma}_{\Lambda kt} - Tr(a_{\Lambda\Sigma kt} B_{\Lambda\Sigma kt}^{-1} \overline{\Sigma}_{\Lambda kt}) - \frac{D}{\beta_{\Lambda\mu kt}})$$

where

$$KL_{Gamma}(a_{vt}, b_{vt}; a_{vt}^{-1}, b_{vt}^{-1}) = (b_{vt} - 1)\Psi_{\gamma}(b_{vt}) - \log a_{vt} - b_{vt} - \log\Gamma_{\gamma}(b_{vt}) + \log\Gamma_{\gamma}(b_{vt}^{-1}) + b_{vt}^{-1}\log(a_{vt}^{-1}) - (b_{vt}^{-1} - 1)(\Psi_{\gamma}(b_{vt}) + \log a_{vt}) + \frac{a_{vt} b_{vt}}{a_{vt}^{-1}}$$

$$KL_{Beta}(\gamma_{vkt1}, \gamma_{vkt2}; \gamma_{vkt1}^{-1}, \gamma_{vkt2}^{-1}) = \ln\left(\frac{B_{\beta}(\gamma_{vkt1}^{-1}, \gamma_{vkt2}^{-1})}{B_{\beta}(\gamma_{vkt1}, \gamma_{vkt2})}\right) + (\gamma_{vkt1} - \gamma_{vkt1}^{-1})\Psi_{\gamma}(\gamma_{vkt1}) + (\gamma_{vkt2} - \gamma_{vkt2}^{-1})\Psi_{\gamma}(\gamma_{vkt2}) + (\gamma_{vkt1}^{-1} - \gamma_{vkt1} + \gamma_{vkt2}^{-1} - \gamma_{vkt2})\Psi_{\gamma}(\gamma_{vkt1} +)$$

$$KL_{W}(a_{\Lambda\Sigma kt}, B_{\Lambda\Sigma kt}; a_{\Lambda\Sigma kt}^{-1}, B_{\Lambda\Sigma kt}^{-1}) = \left(\frac{a_{\Lambda\Sigma kt} - D - 1}{2}\right)\left(\sum_{d=1}^{D}\Psi_{\gamma}\left(\frac{a_{\Lambda\Sigma kt} + 1 - d}{2}\right) - \log\left|B_{\Lambda\Sigma kt}\right| + D\log 2\right) - \left(\frac{a_{\Lambda\Sigma kt}^{-1} - D - 1}{2}\right)\left(\sum_{d=1}^{D}\Psi_{\gamma}\left(\frac{a_{\Lambda\Sigma kt}^{-1} + 1 - d}{2}\right) - \log B_{\Lambda\Sigma kt}^{-1} + D\log 2\right) - \frac{Da_{\Lambda\Sigma kt}}{2} + \frac{a_{\Lambda\Sigma kt}}{2}Tr\left(\frac{B_{\Lambda\Sigma kt}^{-1}}{B_{\Lambda\Sigma kt}}\right) + \log\frac{2^{\frac{Da_{\Lambda\Sigma kt}^{-1}}{2}}\,|B_{\Lambda\Sigma kt}^{-1}|^{\frac{-a_{\Lambda\Sigma kt}^{-1}}{2}}\,\pi^{\frac{D(D-1)}{4}}\prod_{d=1}^{D}\Gamma_{\gamma}\left(\frac{a_{\Lambda\Sigma kt}^{-1} - d + 1}{2}\right)}{2^{\frac{Da_{\Lambda\Sigma kt}}{2}}\,|B_{\Lambda\Sigma kt}|^{\frac{-a_{\Lambda\Sigma kt}}{2}}\,\pi^{\frac{D(D-1)}{4}}\prod_{d=1}^{D}\Gamma_{\gamma}\left(\frac{a_{\Lambda\Sigma kt} - d + 1}{2}\right)}$$

$$KL_{N}\left(m_{\Lambda\mu kt}, \frac{\Sigma_{\Lambda kt}}{\beta_{\Lambda\mu kt}a_{\Lambda\Sigma kt}}; m_{\Lambda\mu kt}^{-1}, \frac{\Sigma_{\Lambda\mu kt}^{-1}}{\beta_{\Lambda\mu kt}^{-1}a_{\Lambda\mu kt}^{-1}}\right) = 0.5\log\frac{\left|\frac{\Sigma_{\Lambda\mu kt}^{-1}}{\beta_{\Lambda\mu kt}^{-1}a_{\Lambda\mu kt}^{-1}}\right|}{\left|\frac{\Sigma_{\Lambda kt}}{\beta_{\Lambda\mu kt}a_{\Lambda\Sigma kt}}\right|} + 0.5Tr\left(\frac{\frac{\Sigma_{\Lambda kt}}{\beta_{\Lambda\mu kt}a_{\Lambda\Sigma kt}}}{\frac{\Sigma_{\Lambda\mu kt}^{-1}}{\beta_{\Lambda\mu kt}^{-1}a_{\Lambda\mu kt}^{-1}}}\right) + 0.5(m_{\Lambda\mu kt} - m_{\Lambda\mu kt}^{-1})^{T}\, inv\left(\frac{\Sigma_{\Lambda\mu kt}^{-1}}{\beta_{\Lambda\mu kt}^{-1}a_{\Lambda\mu kt}^{-1}}\right)(m_{\Lambda\mu kt} - m_{\Lambda\mu kt}^{-1}) - \frac{D}{2}$$

It is important to note that anywhere possible, we use log scaled values to prevent underflow. DP also like iHMM is exchangeable, so it is subject to label switching. In other words, switching the labels of partitions does not change the likelihood or posterior.

## Motivation for Gamification Utility Structure

I start this section with explaining the choices of the gamification platform. In particular the gamification elements that I considered include: fun element, badges, leaderboard, and reputation points. For example, a gamification platform might work on the positive environment of social interaction between content producers and consumers, by putting emphasis on different contents, to make the engagement more fun. It can also manipulate the threshold of earning badges, to make earning badges harder or simpler. In addition, a gamification platform can send empowering messages to users whose rank fall on the leaderboard. To find the effect of each of these policies, the gamification platform should measure the response of the users to the gamification incentives.

In the context of this study the choice of users to create content can be in the following forms: to post an answer, to review, or to comment on a question or an answer, so I considered the outcome of the user choice positive if the user makes any of these choices, and negative if the user selects none. Assuming that a contributor has a random state dependent utility, and that the distribution of the random error term is extreme value, a logit function can model the probability of observing a user contribution. As a result, the likelihood of users multiple contributions, follow binomial distribution. Next I explain the rationale behind the variable that might explain the observed state of the users' utility, in terms of the gamification components.

The proposed model includes user and day fixed effect to capture users' heterogeneous optimal stimulation level and its variation across days, because users require motivation to contribute content (Salcu and Actrinei 2013; Mittelstaedt 1976; Joachimsthaler and Lastovicka 1984; Steenkamp and Baumgartner 1992). To capture the interdependence of users' stimulation level, the model specification includes the same prior on the fixed effect of users within each segment. Further, the same prior for the fixed effect of days considers that emotional stimulation across days have the same mean and variance.

The total cumulative number of contributions acts as proxy for the fun that a user experiences. As a result, a lag cumulative number of contributions might be a state variable to capture the effect of the fun elements of the gamification platform. Furthermore, the number of content received (i.e. answer to the posted question) act as the proxy for the social utility of the user. As a result, I included the lagged total number of answered reviewed, and answer accepted by a user, as a proxy for the users' reciprocity state. Another proxy for the social utility of users to contribute content is the level of reputation points, i.e. the number of up-votes a user has received (Bolton et al. 2013; Bolton et al. 2004; Yoganarasimhan 2013; Lee and Bell 2013; Toubia and Stephen 2013). As the reputation point might have both instant and long term effects, the utility of the user incorporates both the weekly level, and the cumulative level of user reputation (Wei et al. 2015; Li et al. 2015). Another gamification element that is proxy signal for social status of user is the lagged leaderboard absolute rank and rank change. The latter one might be relevant for potential endowment effect. In other words, an individual might be regretful for losing the last week rank or forgone social status.

Last but not least, badges might also affect users' motivations to contribute content, for both intrinsic (empowerment effect), or extrinsic (social status function) motivations (e.g. Antin and Churchill 2011; Wei et al. 2015; Li et al. 2015). Two types of variables capture the effect of badges: badge category (i.e. Gold, Silver, Bronze), and tagged badge category (i.e. knowledge domain tag of the badge). A user might have different preferences for different tag badges. For example, a user might value badge of gold contributor to R programming community tag more than badge of gold contributor to C++ community tag, because he wants to build reputation as a data scientist. I allowed for heterogeneity in the tag badge effects. In addition, a gold badge in any community might have its own value, for creating gold member status. Furthermore, Gold, Bronze, and Silver define different game levels. In summary, I captured the effect of gold, bronze, and silver status of the badges in the badge category variable, and the effect of getting either of these tagged badges, as a proxy for progress in different knowledge domains.

Like the effect of any marketing policy, short term and long term effect of earning the badges might be different (Liu 2007;Jedidi et al. 1999, Mela et al. 1997; Lewis 2004). As a result, consistent with Wei et al. (2015) and Li et al. (2015), the utility of consumers includes both lagged cumulative and instant number of each of the badges. Figure 3.4 shows box and arrow diagram of the components of the state-dependent utility of users to contribute content.

Figure 3.1. Box and Arrow Model of State Dependent Utility of a user to contribute

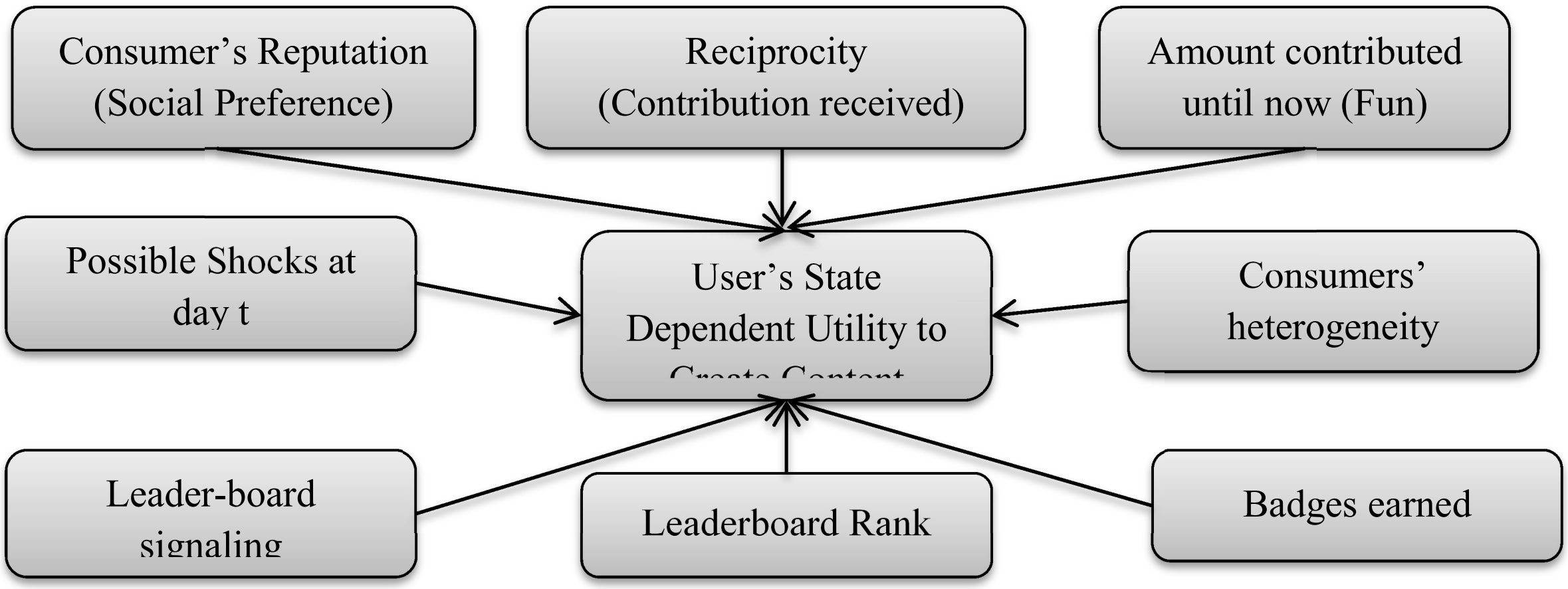

Last but not least, to control for the heterogeneity in users' responses to each of the gamification elements, the model of the users' state-dependent utility allows for flexible patterns of response, through a random coefficient model. Formally, the users' state-dependent utility of user i, in segment c, at day t, in week w, has the following form:

$$\begin{aligned} U_{ict} = {} & \alpha_i + \beta_t + \gamma_{c1} cont_{it-1} + \gamma_{c2} rcv_{it-1} + \gamma_{c3} crep_{iw-1} + \gamma_{c4} rep_{iw-1} + \\ & \gamma_{c5} rnk_{iw-1} + \gamma_{c6} \Delta rnk_{iw-1} + \gamma_{c7} bdg_{it-1} + \gamma_8 tag_{it-1} + \\ & \gamma_{c9} cbdg_{it-1} + \gamma_{10} ctag_{it-1} + \varepsilon_{ict} \end{aligned} \quad (1)$$

Table 3.6 summarizes the definition of each of the variables in the model.

Table 3.1. Model Variables Definition

| Variable | Description |
|---|---|
| State Dependent Utility( $U_{ict}$ ) | State dependent utility of Consumer i in segment c at time t |
| Individual Specific Fixed Effect $(\alpha_i)$ | Fixed effect, or fixed optimal threshold level of individual i |
| Day Fixed Effect $(\beta_t)$ | Fixed effect of day t, or average effect of day t on the optimal motivation level |
| Contribution State ( $cont_{it-1}$ ) | Total contribution level of individual i, up until the current contribution point in time, demeaned and then normalized by a hundred |
| Reciprocity State ( $rcv_{it-1}$ ) | Total number of contribution received by individual i, up until day t, demeaned and normalized by a hundred |
| Reputation State ( $crep_{iw-1}$ ) | Total number of reputation points received by individual i, up until week w |
| Weekly Reputation ( $rep_{iw-1}$ ) | Total number of reputation point received by individual i, at the previous week (i.e. week w-1) |
| Leaderboard rank ( $rnk_{iw-1}$ ) | Rank of individual i, in the leaderboard at previous week (i.e. week w-1) |
| Leaderboard rank change ( $\Delta rnk_{iw-1}$ ) | Change in the individual i's rank in the leaderboard from the other week to the previous week (i.e. week w-2 to week w-1) |
| Instant Badge category ( $bdg_{it-1}$ ) | A vector of number of gold, silver, and bronze badges individual i earned at the previous day (i.e. day t-1) |
| Instant Tag Badges ( $tag_{it-1}$ ) | A vector of the number of badges pertained to each of the badge categories that individual i earned until the previous day (i.e. day t-1) |
| Cumulative Badge Category ( $cbdg_{it-1}$ ) | A vector of total cumulative number of gold, silver, and bronze badges individual i earned until the previous day (i.e. day t-1) |
| Cumulative Tag Badges ( $ctag_{it-1}$ ) | A vector of the total cumulative number of badges pertained to each of the badge categories that individual i earned until the previous day (i.e. day t-1) |
| $\gamma_{c1},\gamma_{c2},\gamma_{c3},\gamma_{c4},\gamma_{c5},\gamma_{c6},\gamma_{c7},\gamma_{c9}$ | Segment c specific parameters of state dependent utility of consumer i in segment c |
| $\gamma_8,\gamma_{10},\varepsilon_{ict}$ | Segment independent parameters of response to tag badges, and type one extreme value error |

To capture this heterogeneity, I used a two-step approach. First, based on the users' cross sectional information presented in table 3.3 and 3.4, i.e. $x_i$, I clustered the users, with mixture normal assumption. Second, I considered that users within each cluster c response differently to each of the state variables defined above. Formally, I used the following specification to cluster the users:

$$f(x_i)=\sum_{c=1}^{C}\pi_c f(x_i \mid \mu_c,\Sigma_c), \sum_c \pi_c = 1$$

Where f is the normal distribution density function, and $(\mu_c,\Sigma_c,\pi_c)$ are the mean, variance, and segment size parameters of each of the segments. Finally, to explain the heterogeneity in the parameters across segments, I use a step wise regression of the parameters of segments on the observed information of each segment, to deal with potential multi-collinearity between large numbers of variables in the observed information vector of each segment.

It might be relevant to note that for computational tractability over a big data set, and parsimony, I defined the model very simple. According to the machine learning anecdotal evidence data always wins over the complex models[1]. In other word, although from modeling perspective, it is possible to include latent motivation levels, and forward looking behavior, such modeling choices not only might make strong assumption about the underlying behavior of consumer in an emotionally laden gamification environment, but also might make the estimation of such model over a big data set intractable.

### References

Beal, M. J., Ghahramani, Z., & Rasmussen, C. E. (2001). The infinite hidden Markov model. In *Advances in neural information processing systems* (pp. 577-584).

Blei, D. M., & Jordan, M. I. (2006). Variational inference for Dirichlet process mixtures. *Bayesian analysis*, *1*(1), 121-143.

Burda, M., Harding, M., & Hausman, J. (2008). A Bayesian mixed logit–probit model for multinomial choice. *Journal of Econometrics*, *147*(2), 232-246.

Carvalho, C., Johannes, M. S., Lopes, H. F., & Polson, N. (2010). Particle learning and smoothing. *Statistical Science*, *25*(1), 88-106.

Carvalho, C. M., Lopes, H. F., & Polson, N. (2009). *Particle learning for generalized dynamic conditionally linear models*. Working paper, University of Chicago Booth School of Business.

Frühwirth-Schnatter, S., & Frühwirth, R. (2007). Auxiliary mixture sampling with applications to logistic models. *Computational Statistics & Data Analysis*, *51*(7), 3509-3528.

Gershman, S. J., & Blei, D. M. (2012). A tutorial on Bayesian nonparametric models. *Journal of Mathematical Psychology*, *56*(1), 1-12.

Ghahramani, Z., & Beal, M. J. (2001). Propagation algorithms for variational Bayesian learning. *Advances in neural information processing systems*, 507-513.

[1] Garrett, W. Why more data and simple algorithms beat complex analytics models. Data informed website. http://data-informed.com/why-more-data-and-simple-algorithms-beat-complex-analytics-models/. August 7, 2013. Accessed June 5, 2015.

Penny, W. D. (2001). Variational Bayes for d-dimensional Gaussian mixture models. *University College London*.

Penny, W.D. “Kullback-Liebler Divergences of Normal, Gamma, Dirichlet and Wishart Densities.” Technical report, Wellcome Department of Cognitive Neurology, 2001.

Netzer, O., Lattin, J. M., & Srinivasan, V. (2008). A hidden Markov model of customer relationship dynamics. *Marketing Science*, *27*(2), 185-204.

Rodriguez, A. (2011). On-line learning for the infinite hidden Markov model.*Communications in Statistics-Simulation and Computation*, *40*(6), 879-893.

Teh, Y. W., Jordan, M. I., Beal, M. J., & Blei, D. M. (2006). Hierarchical dirichlet processes. *Journal of the american statistical association*, *101*(476).

Van Gael, J., Saatci, Y., Teh, Y. W., & Ghahramani, Z. (2008, July). Beam sampling for the infinite hidden Markov model. In *Proceedings of the 25th international conference on Machine learning* (pp. 1088-1095). ACM.

Wood, F., & Black, M. J. (2008). A nonparametric Bayesian alternative to spike sorting. *Journal of neuroscience methods*, *173*(1), 1-12.